\documentclass[sigconf]{acmart}

\usepackage{amsthm}
\usepackage{amsmath}
\usepackage{multirow}
\usepackage{color}
\usepackage{colortbl}
\usepackage{xcolor}
\definecolor{myred}{HTML}{F8CECC}
\definecolor{myyellow}{HTML}{FFF2CC}
\definecolor{myblue}{HTML}{DAE8FC}

\newcommand{\yellow}[1]{\colorbox{myyellow!80}{#1}}
\newcommand{\red}[1]{\colorbox{myred!80}{#1}}
\newcommand{\blue}[1]{\colorbox{myblue!80}{#1}}

\DeclareMathOperator*{\argmax}{arg\,max}

\newcommand{\repourl}{https://github.com/waylonli/TSPRank-KDD2025}

\makeatletter
  \if@ACM@anonymous
    \renewcommand{\repourl}{https://anonymous.4open.science/r/TSPRank-Submission-583F/}
  \fi
\makeatother

\newenvironment{itemizesquish}[2]{\begin{list}{\labelitemi}{\setlength{\itemsep}{#1}\setlength{\labelwidth}{#2}\setlength{\leftmargin}{\labelwidth}\addtolength{\leftmargin}{\labelsep}}}{\end{list}}

\AtBeginDocument{%
  }

\copyrightyear{2025}
\acmYear{2025}
\setcopyright{rightsretained}
\acmConference[KDD '25]{Proceedings of the 31st ACM SIGKDD Conference on Knowledge Discovery and Data Mining V.1}{August 3--7, 2025}{Toronto, ON, Canada} 
\acmBooktitle{Proceedings of the 31st ACM SIGKDD Conference on Knowledge Discovery and Data Mining V.1 (KDD '25), August 3--7, 2025, Toronto, ON, Canada} 
\acmDOI{10.1145/3690624.3709234} 
\acmISBN{979-8-4007-1245-6/25/08}

\begin{document}

%%
%% The "title" command has an optional parameter,
%% allowing the author to define a "short title" to be used in page headers.
\title[TSPRank: Bridging Pairwise and Listwise Methods with a Bilinear Travelling Salesman Model]{TSPRank: Bridging Pairwise and Listwise Methods with a Bilinear Travelling Salesman Model}
% TODO mixed integer key contributions embedding ranking etc.

%%
%% The "author" command and its associated commands are used to define
%% the authors and their affiliations.
%% Of note is the shared affiliation of the first two authors, and the
%% "authornote" and "authornotemark" commands
%% used to denote shared contribution to the research.
\author{Weixian Waylon Li}
\affiliation{%
  \institution{University of Edinburgh}
  % \city{Edinburgh}
  \country{United Kingdom}
}
\email{waylon.li@ed.ac.uk}

\author{Yftah Ziser}
\affiliation{%
  \institution{Nvidia Research}
  % \city{}
  \country{Israel}}
\email{yziser@nvidia.com}

\author{Yifei Xie}
\affiliation{%
  \institution{University of Edinburgh}
  % \city{Edinburgh}
  \country{United Kingdom}
}
\email{yifei.xie@ed.ac.uk}

\author{Shay B. Cohen}
\affiliation{%
  \institution{University of Edinburgh}
  % \city{Edinburgh}
  \country{United Kingdom}
}
\email{scohen@inf.ed.ac.uk}

\author{Tiejun Ma}
\affiliation{%
  \institution{University of Edinburgh}
  % \city{Edinburgh}
  \country{United Kingdom}
}
\email{tiejun.ma@ed.ac.uk}

%%
%% By default, the full list of authors will be used in the page
%% headers. Often, this list is too long, and will overlap
%% other information printed in the page headers. This command allows
%% the author to define a more concise list
%% of authors' names for this purpose.

%%
%% The abstract is a short summary of the work to be presented in the
%% article.
\begin{abstract}
% Learning to Rank (LETOR) algorithms are crucial in applications like recommendation systems, question answering, information retrieval, and ranking tasks in other domains such as stock ranking and textual ordering. 
Traditional Learning-To-Rank (LETOR) approaches, including pairwise methods like RankNet and LambdaMART, often fall short by solely focusing on pairwise comparisons, leading to sub-optimal global rankings. 
Conversely, deep learning based listwise methods, while aiming to optimise entire lists, require complex tuning and yield only marginal improvements over robust pairwise models.
To overcome these limitations, we introduce Travelling Salesman Problem Rank (TSPRank), a hybrid pairwise-listwise ranking method. 
TSPRank reframes the ranking problem as a Travelling Salesman Problem (TSP), a well-known combinatorial optimisation challenge that has been extensively studied for its numerous solution algorithms and applications. 
This approach enables the modelling of pairwise relationships and leverages combinatorial optimisation to determine the listwise ranking.
This approach can be directly integrated as an additional component into embeddings generated by existing backbone models to enhance ranking performance.
Our extensive experiments across three backbone models on diverse tasks, including stock ranking, information retrieval, and historical events ordering, demonstrate that TSPRank significantly outperforms both pure pairwise and listwise methods.
Our qualitative analysis reveals that TSPRank's main advantage over existing methods is its ability to harness global information better while ranking.
TSPRank's robustness and superior performance across different domains highlight its potential as a versatile and effective LETOR solution.
The code and preprocessed data are available at \url{\repourl}.
\end{abstract}

%%
%% The code below is generated by the tool at http://dl.acm.org/ccs.cfm.
%% Please copy and paste the code instead of the example below.
%%


%%
%% Keywords. The author(s) should pick words that accurately describe
%% the work being presented. Separate the keywords with commas.
%% A "teaser" image appears between the author and affiliation
%% information and the body of the document, and typically spans the
%% page.

% \received{20 February 2024}
% \received[revised]{12 March 2009}
% \received[accepted]{5 June 2009}

%%
%% This command processes the author and affiliation and title
%% information and builds the first part of the formatted document.
\maketitle

% Page Limit: 8 pages excluding reference and appendix

\section{Introduction}
\label{sec:introduction}

Learning to Rank (LETOR) algorithms have become essential in applications such as recommendation systems \cite{ada-ranker,peiPersonalizedRerankingRecommendation2019,rank-insights}, question answering \cite{10.1145/2505515.2505670,campese-etal-2024-pre}, and information retrieval \cite{aiLearningDeepListwise2018a,pang2020setrank,rankformer}. 
These algorithms aim to order a list of ranking entities based on their features, optimising for the most relevant or preferred entities to appear at the top. 
Recently, LETOR methods have also expanded into other domains, such as stock and portfolio selection \cite{SONG201720,portfolioselectionletor} and textual ordering \cite{tan-etal-2013-learning,NEURIPS2023_f2a11632}.
Despite these broader applications of LETOR, latest fundamental work on LETOR still primarily focused on incorporating click data into ranking models and addressing biases introduced by user feedback \cite{10.1145/3569453,10.1145/3580305.3599914,10.1145/3394486.3403336}, concentrating on retrieval and recommendation tasks. 
Limited research has explored general ranking methods across diverse tasks and domains.

Over the past decade, research on LETOR models mainly focuses on pairwise and listwise approaches, while pointwise methods often fail to capture the intricate inter-entity relationships that are essential for accurate ranking. 
Pairwise methods, such as LambdaMART and RankNet  \cite{lambdamart}, primarily optimise for pairwise comparisons without a holistic view of the entire ranking list, potentially leading to sub-optimal global rankings. 
Listwise methods optimise the ranking of entire lists directly rather than individuals or pairs. 
This category includes advanced neural network architectures, particularly adaptations of the Transformer architecture \cite{attention}, such as Rankformer \cite{rankformer} and SetRank \cite{pang2020setrank}.
However, existing work shows that deep learning-based listwise models require complex tuning to achieve marginal gains over robust pairwise models like LambdaMART on information retrieval benchmarks \cite{google-neural-outperform-gbdt}.
Therefore, pairwise and listwise methods each have inherent drawbacks and more effective solutions for LETOR have not been exhaustively explored.

Recently, GNNRank successfully used directed graph neural networks to learn listwise rankings from pairwise comparisons \cite{he2022gnnrank}. 
Although their method is not applicable to general LETOR problems, as it requires pre-known pairwise relationships like the outcomes of sports matches, it still suggests a potential approach for LETOR by representing pairwise comparisons as a graph.
Inspired by \citeauthor{abend-etal-2015-lexical}, who used combinatorial optimisation techniques to recover the lexical order of recipe pieces \cite{abend-etal-2015-lexical}, we propose Travelling Salesman Problem Rank (TSPRank). 
This hybrid pairwise-listwise method reframes the ranking problem as a Travelling Salesman Problem (TSP), an NP-hard combinatorial optimisation challenge that seeks the optimal sequence of node visits to minimise total travel cost. 
In a ranking context, TSPRank aims to find the optimal permutation of ranking entities to maximise the ranking score.

% Narrow down the claims
Our main contribution, TSPRank, bridges the gap between pairwise and listwise ranking models through combinatorial optimisation, demonstrating strong robustness across diverse ranking problems. 
TSPRank simplifies the complex listwise ranking problem into easier pairwise comparisons, thus overcoming the challenges of directly learning listwise rankings in complex data and tasks. 
It also enhances pairwise methods by incorporating listwise optimisation through a TSP solver. To improve ranking performance, TSPRank can be applied directly as an additional component on embeddings generated from domain-specific backbone models. 
To our knowledge, it is the first LETOR model to frame the ranking problem as a combinatorial optimisation plus graph representation task. 
We demonstrate its superior performance across three backbone models on diverse ranking tasks: stock ranking, information retrieval, and historical events ordering, covering both numerical and textual data across multiple domains. We introduce two learning methods for TSPRank: a local method using a pre-defined ground truth adjacency matrix without the TSP solver during training and a global end-to-end method that includes the solver in the training loop for better model-inference alignment. Our empirical analysis visualises the predictions and graph connections to provide insights into TSPRank’s effectiveness. 
We demonstrate that the TSP solver enhances the model’s ability to harness global information during ranking and increases tolerance to errors in pairwise comparisons. 
Additionally, we assess inference latency caused by the combinatorial optimisation solver and suggest potential solutions to mitigate this overhead.

% \shaycomment{there seems to be no related work section, do we need one for this conference? or did you mean to include that in Background?}
% \waylon{I have changed the section title to related work. I was thinking about including it in Background but I did find that most of the accepted papers do contain the Related Work section.}

\section{Background}

% \subsection{Learning To Rank}
\label{sec:letor}
LETOR algorithms aim to rank entities based on their features. %It has been successfully applied in many areas such as recommendation~\cite{duan2010empirical,ada-ranker,rank-insights}, question answering~\cite{yang2016beyond, reimers2019sentence}, and retrieval~\cite{joachims2002optimizing, liu2009learning,10.1145/3477495.3531958,10.1145/3477495.3531948}. %% repetitive
In LETOR, we denote a list of $N$ ranking entities as $\mathbf{e}$, containing $\{\mathbf{e}_1, \mathbf{e}_2, \dots, \mathbf{e}_N\}$, with each $\mathbf{e}_i \in \mathbb{R}^d$ where $d$ represents the dimensionality of the feature space. A scoring function $s: \mathbb{R}^{d} \rightarrow \mathbb{R}$ is applied on the entities, which are then ranked in descending order of the $s(\mathbf{e}_i)$ scores. LETOR algorithms are broadly categorised into pointwise, pairwise, and listwise approaches based on their optimisation strategies. 
Pointwise methods treat ranking as a regression problem, comparing scores $s(\mathbf{e}_i)$ to labels $y_i$. 
Models such as OPRF~\cite{fuhr1989optimum}, TreeBoost~\cite{friedman2001greedy}, and RankSVM~\cite{shashua2002taxonomy}, including newer implementations like RankCNN~\cite{severyn2015learning}, are known for their computational efficiency but may not capture inter-entity relationships adequately.
% \shaycomment{I think what you mean by pointwise and listwise is related to local/global learning, etc. - mention it in that context. also I don't quite understand how we differ from either - are we pointwise or listwise, in which sense we are either? clarify.}
% \waylon{We are using a pairwise-listwise method. Pointwise simply means modelling a ranking problem as a regression problem, only considering the datapoints in isolation. What we do is encode the pairwise relationship and do listwise inference, no matter local or global learning.}

Pairwise methods focus on the relative comparisons of entity pairs, assessing scores $(s(\mathbf{e}_i), s(\mathbf{e}_j))$ against binary labels $y_{ij}$, which denote if entity $\mathbf{e}_i$ should be ranked higher than $\mathbf{e}_j$.
Methods in this category include RankNet~\cite{burges2005learning}, LambdaRank~\cite{burges2006learning}, and LambdaMART~\cite{lambdamart}. Listwise methods directly compute scores based on the features of all entities in a list. 
These methods aim to optimise a listwise objective, comparing the entire set of computed scores $s(\mathbf{e}) \in \mathbb{R}^N$ against a complete set of labels $\mathbf{y} \in \mathbb{R}^N$. 
To advance this approach, \citeauthor{aiLearningDeepListwise2018a} introduced the Deep Listwise Context Model (DLCM) \cite{aiLearningDeepListwise2018a}. This model was further refined into general multivariate scoring functions in \cite{aiLearningGroupwiseMultivariate2019}.
Considering that a list of entities can be viewed as a sequence, methodologies from Natural Language Processing (NLP) are applicable to ranking tasks. 
Notably, the Transformer architecture \cite{attention}, has been adapted to address listwise ranking challenges, referred to as listwise Transformers in recent work \cite{peiPersonalizedRerankingRecommendation2019, pasumarthiPermutationEquivariantDocument2020, pang2020setrank, zhuangCrossPositionalAttentionDebiasing2021, liPEARPersonalizedReranking2022, rankformer}.

Although listwise ranking methods, primarily neural rankers, have demonstrated their advantages in web search reranking tasks \cite{aiLearningDeepListwise2018a, pang2020setrank}, \citeauthor{google-neural-outperform-gbdt} conducted a comprehensive evaluation comparing listwise neural rankers to the gradient-boosted-decision-tree (GBDT)-based pairwise LambdaMART \cite{lambdamart,google-neural-outperform-gbdt}. 
The results indicated that LambdaMART consistently outperforms listwise rankers across three widely used information retrieval datasets. Furthermore, it was observed that listwise rankers require substantial efforts to achieve only marginal improvements over LambdaMART. 
Consequently, it is evident that listwise ranking models are not universally applicable solutions and generally lack the robustness needed for broader deployment.
This conclusion has also been further validated by \citeauthor{rankformer} \cite{rankformer}.

\section{TSPRank}
\label{sec:tsprank}

% Apart from the identified limitations of neural ranking models, such as issues with feature transformation, network architecture, and data sparsity noted in previous work \cite{google-neural-outperform-gbdt}, 
Directly predicting the listwise order is challenging, as correctly ranking $N$ entities from 1 to $N$ is complex. 
However, breaking it down into $N \times N$ pairwise comparisons simplifies the task, as each pairwise comparison is more straightforward than the entire listwise ranking. 
However, a fundamental drawback of pairwise ranking, as highlighted by \citeauthor{pairwise-to-listwise} \cite{pairwise-to-listwise}, is that the learning objectives do not focus on minimising the listwise ranking errors. 
To address this gap and rethink the approach of pairwise ranking, we introduce TSPRank, a novel methodology for position-based ranking problems. 
This approach integrates listwise optimisation into pairwise comparisons by modelling ranking tasks as TSP. 
Notably, TSPRank can be applied as an additional component with embeddings generated from domain-specific backbone models to achieve improved ranking performance.

\subsection{Ranking As Travelling Salesman Problem}
\label{sec:formulation-tsp}

% The Travelling Salesman Problem (TSP) in the context of ranking is defined as follow: Given a fully connected graph consisting of nodes $\{n_1, n_2, ..., n_N\}$ where each node represents an entity to be ranked, and edges between each pair of nodes weighted by a score $s(n_i, n_j)$ that quantifies the gain of placing node $n_j$ immediately after node $n_i$ in a ranking. The objective is to find a permutation $\pi$ of these nodes that results in the maximum total score, effectively ranking the entities in an optimal sequence with respect to the given scores. The mathematical formulation of this problem is to explicitly seek the particular permutation $\pi^*$ that results in the highest total transition score, defined as follows:

We model the ranking problem as an open-loop Travelling Salesman Problem (TSP). Given a set of entities ${\mathbf{e}_1, \mathbf{e}_2, \ldots, \mathbf{e}_N}$ and all the pairwise score values $s(\mathbf{e}_i, \mathbf{e}_j)$ indicating the gain of ranking entity $\mathbf{e}_j$ immediately after $\mathbf{e}_i$, we define a complete graph $G = (V, W)$, where $V$ represents the entities and $W$ represents the pairwise scores between every two entities. 
Each entity $\mathbf{e}_i$ in the ranking problem corresponds to a city in the TSP.
By mapping entities to cities and pairwise scores to weights, we can formulate the ranking problem as a TSP to find the optimal permutation $\pi$. Thus, the objective is to find an optimal open-loop tour $\pi^*$ of the entities defined in Equation~\ref{eq:op-prob} such that the total pairwise score is maximised.

\begin{equation}
\label{eq:op-prob}
     \pi^* = \underset{\pi}{\mathrm{argmax}} \sum_{i=1}^{N-1} s(\mathbf{e}_{\pi(i)}, \mathbf{e}_{\pi(i+1)})\text{.}
\end{equation}

The next step is developing a scoring function that accurately models the pairwise relationships between the ranking entities, thereby effectively determining the scores.

\subsection{Scoring Model}

The scoring model takes a set of ranking entities or their corresponding embeddings $\{\mathbf{e}_1, \mathbf{e}_2, \dots, \mathbf{e}_N\}$, where each $\mathbf{e}_i \in \mathbb{R}^d$ and $d$ is the dimension of the features or embeddings. 
It outputs an $N \times N$ weighted adjacency matrix $A$ that represents the pairwise scores of every two entities. 
The scoring model comprises an optional node encoder and a trainable bilinear model on top.

\paragraph{Encoder}
The default encoder used is the Transformer encoder block \cite{attention}, but it can be replaced with any other encoder suited to the specific task and dataset. 
Additionally, using an encoder is optional if sufficiently robust backbone models are available.

\paragraph{Trainable Bilinear Model}
Given a pair of encoded representations of entities $(\mathbf{e}_i, \mathbf{e}_j)$, the bilinear model computes the pairwise score using the following bilinear form:

\begin{equation}
    s(\mathbf{e}_i, \mathbf{e}_j) = \mathbf{e}_i^\intercal \mathbf{W} \mathbf{e}_j + b,
\end{equation}

\noindent where $\mathbf{W}$ and $b$ are learnable parameters that can be optimised along with the encoder parameters.

The adjacency matrix $A$ is constructed as follows:

\begin{equation}
    A_{ij} = s(\mathbf{e}_i, \mathbf{e}_j), \  \text{for all } i, j \in \{1, \dots, N\},
\end{equation}

\noindent where each entry $A_{ij}$ directly corresponds to the computed pairwise comparison score from entity $\mathbf{e}_i$ to entity $\mathbf{e}_j$.

\subsection{Inference}

Inference of TSPRank, described in Equation~\ref{eq:op-prob}, requires solving the open-loop TSP, which is widely recognised as an NP-hard combinatorial optimisation problem. 
The solution algorithms of the TSP consist of exact algorithms (brunch and bound \cite{poikonen2019branch}) and approximation algorithms (ant colony optimisation \cite{yang2008ant}).
Given the scale of the ranking problem and the desired ranking accuracy, we prefer to obtain an exact solution for the TSP rather than an approximate one. 
Therefore, we formulate the TSP as a Mixed Integer Linear Programming (MILP) problem \cite{rieck2010new}, where the integer variables represent binary decisions on whether a path between two nodes is included in the optimal tour. 
We then use an MILP optimization solver, such as CPLEX \cite{cplex} or Gurobi \cite{gurobi}, to find the exact optimal solution.

The decision variables are defined as follows: 
\begin{equation}
\label{eq:decision-var}
    x_{ij} = 
        \begin{cases}
            1,& \text{if entity $\mathbf{e}_j$ is ranked immediately after $\mathbf{e}_i$.} \\
            0,              & \text{otherwise.}
    \end{cases}
\end{equation}

We then apply the TSP optimisation problem defined in Appendix~\ref{appendix:tsp-constraints} to ensure the result is a single, complete ranking that includes all entities.
This inference process effectively performs listwise optimisation, as it determines the optimal permutation of entities that maximises the overall pairwise ranking score.

\section{Learning}
\label{sec:learning}
\begin{figure*}[htbp]
    \centering
    \includegraphics[width=0.95\textwidth]{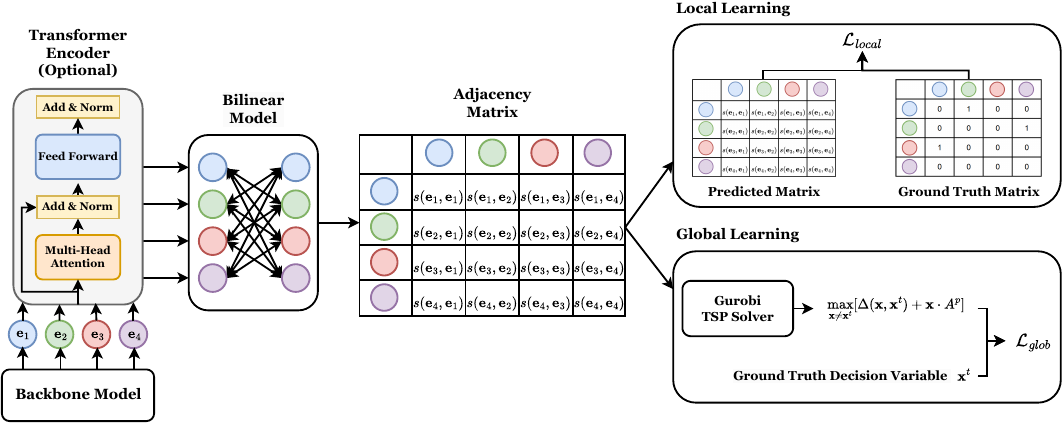}
    \caption{Illustration of TSPRank and the complete pipeline of local and global learning. The pipeline starts with a Transformer Encoder or any embeddings generated from another existing Backbone Model, followed by a Bilinear Model generating pairwise scores to form an Adjacency Matrix. Local learning compares the predicted matrix with the ground truth to calculate the local loss $\mathcal{L}_{local}$. Global learning uses the max-margin loss $\mathcal{L}_{glob}$ to incorporate the Gurobi TSP solver during training.}
    \label{fig:tsprank-arch}
    \vspace{-0.2cm}
\end{figure*}

We introduce two distinct learning methodologies for the TSPRank model outlined in Section~\ref{sec:tsprank}.
The first method is a local approach, focusing exclusively on individual pairwise comparisons during the training phase. 
The second method is a global, end-to-end approach, incorporating the Gurobi solver directly into the training process, thereby allowing the black-box MILP solver to influence the learning dynamically \cite{li-etal-2023-bert}.
Figure~\ref{fig:tsprank-arch} illustrates the complete pipeline of the two learning methods as well as the model architecture.

\subsection{Local Learning}
\label{sec:local-learning}

In the local learning approach, the objective is to understand pairwise consecutive relationships between nodes.
The objective is to determine whether entity $\mathbf{e}_j$ should rank one position after $\mathbf{e}_i$ in a given pair of nodes. Therefore, the score $s(\mathbf{e}_i, \mathbf{e}_j)$ is set high if $\mathbf{e}_j$ is supposed to rank consecutively after $\mathbf{e}_i$, and low otherwise.

To tailor the TSPRank model to ranking scenarios where penalties vary based on the actual positions, we apply a weighted loss. 
We denote the predicted adjacency matrix $A^p$ from the bilinear model, where $a_{ij}^p \in A^p$ represents the predicted pairwise scores. 
The target adjacency matrix $A^t = \{a^t_{ij}\}$ is defined such that $a_{ij}^t \in A^t$ equals 1 if $\mathbf{e}_j$ ranks immediately after $\mathbf{e}_i$, and 0 otherwise. 
We apply the weighted cross-entropy loss (defined in Equation~\ref{eq:entropy-loss}) on each row to identify the next entity $\mathbf{e}_{\pi(i+1)}$ to be ranked given $\mathbf{e}_{\pi(i)}$.
The model is trained as a multi-class classification problem, aiming to maximise the probability of $P(\mathbf{e}_{\pi(i+1)} \mid \mathbf{e}_{\pi(i)})$.
We weight the loss by the true ranking label $y$, allowing the penalty to vary according to different ranking positions. 
Additionally, $y$ can be adjusted to $N + 1 - y$ depending on whether $y$ represents ascending or descending order.

\begin{equation}
\label{eq:entropy-loss}
\mathcal{L}_{local}(A^p, A^t) = - \sum_{i=1}^N y_k \log \frac{e^{A^p_{ik}}}{\sum_{j=1}^N e^{A^p_{ij}}},\  k = \argmax_j A^t_{ij}.
\end{equation}

% \begin{equation}
%     \mathcal{L}(A^p, A^t) = \sum_{i=1}^N \sum_{j=1}^N - \frac{w^{g}_{ij}}{0.5 \cdot N} [ a^t_{ij} \cdot \log a^p_{ij} + (1-a^t_{ij}) \cdot \log (1 - a^p_{ij})]
% \end{equation}

% The weighting matrix $\mathbf{w}^g$ is defined by the true rank difference gap:

% \begin{equation}
% \mathbf{w}^g =
%     \begin{bmatrix}
%         0 & |g_1 - g_2| & \dots & |g_1 - g_N| \\
%         |g_2 - g_1| & 0 & \dots & |g_2 - g_N| \\
%         \vdots & \vdots & \ddots & \vdots \\
%         |g_N - g_1| & |g_N - g_2| & \dots & 0
%     \end{bmatrix},
% \end{equation}

% where $g_1, \dots, g_N$ are the gold standard rankings of the entities.

\subsection{Global Learning}

We expect that a globally trained TSPRank model will further enhance the performance as the local learning method focuses solely on pairwise comparisons in isolation during training. 
Incorporating the TSP solver in the training procedure will better align the model with the inference process.
Therefore, we introduce an end-to-end approach. 
After obtaining the predicted adjacency matrix $A^p$, defined in Section~\ref{sec:local-learning}, we define the gold decision variable $\mathbf{x}^t = \{x^t_{ij}\}$, where $x^t_{ij}$ is defined as in Equation~\ref{eq:decision-var}. 
We aim to train a model to satisfy the margin constraints (Equation~\ref{eq:margin-cons}) for all the possible decision variables $\mathbf{x}$ by minimising the max-margin loss defined in Equation~\ref{eq:max-margin-loss}.
When $\mathbf{x} = \mathbf{x}_t$, the predicted ranking perfectly aligns with the gold ranking, resulting in a loss of zero.

\begin{align}
\label{eq:margin-cons}
    \mathbf{x}^t \cdot A^p & \ge \mathbf{x} \cdot A^p + \Delta (\mathbf{x}^t, \mathbf{x}), \text{ for all } \mathbf{x} \text{,}\\
\label{eq:max-margin-loss}
    \mathcal{L}_{glob} (A^p, \mathbf{x}^t) & = \max (0, \max_{\mathbf{x} \neq \mathbf{x}^t} [\Delta (\mathbf{x}, \mathbf{x}^t) + \mathbf{x} \cdot A^p ] - \mathbf{x}^t \cdot A^p )
\end{align}

The structured score $\Delta (\mathbf{x}, \mathbf{x}^t)$, defined in Equation~\ref{eq:structured-score}, aims to enforce a margin for each incorrectly identified edge.

\begin{equation}
\label{eq:structured-score}
    \Delta (\mathbf{x}, \mathbf{x}^t) = \sum_{i=1}^N \sum_{j=1}^N \max (0, x_{ij} - x_{ij}^t)\text{.}
\end{equation}

This approach eliminates the need to manually define the target adjacency matrix.
Instead, it uses the output from the black-box optimiser to directly guide the updates of the model parameters. 
This methodology not only simplifies the process but also ensures a closer alignment with the solver's procedural dynamics.
To accelerate convergence, we integrate a hybrid of local and global loss. Relying solely on the global loss requires significantly more epochs to converge. Therefore, during training, we alternate batches between the global loss and the local loss, ensuring faster and more efficient model convergence.

\section{Experimental Setup}
\label{sec:exp-setup}

We evaluate TSPRank’s effectiveness as a prediction layer integrated into various backbone models. 
Our goal is to demonstrate that TSPRank enhances performance compared to the original backbone and other general ranking algorithms across different domains. While achieving state-of-the-art (SOTA) performance still depends on the backbone design, TSPRank serves as a component to improve ranking performance. 
We focus on scenarios where (i) the data contain ordinal ranking labels rather than binary or relevance level labels, and (ii) the complete ranking or at least the top-k entities are important, extending the task’s interest beyond merely identifying the top-1 entity. 
We test our method with three datasets and backbone models from diverse domains: stock ranking, information retrieval, and historical events ordering. 
These datasets cover various data modalities, including tabular and textual data. 
All three tasks can be formulated as presented in Section~\ref{sec:formulation-tsp}. 
We aim to answer the following research questions (\textbf{RQs}):

\begin{itemizesquish}{0em}{0em}
\item \textbf{RQ1}: Does the pairwise-listwise TSPRank outperform the current SOTA general pairwise and listwise ranking methods across different backbone models and datasets?
\item \textbf{RQ2}: Does global learning lead to better performance?
\item \textbf{RQ3}: Does the pure pairwise method really consistently outperform the deep learning based listwise method across different domains?
\item \textbf{RQ4}: What are the advantages of hybrid pairwise-listwise TSPRank over other methods?
\end{itemizesquish}

This section details the experimental setup for each task. All experiments were conducted on a single Nvidia A100 80G GPU.

\subsection{Benchmark Models}

We benchmark our pairwise-listwise TSPRank model against the original backbone models (if they originally contain a prediction layer), LambdaMART \cite{lambdamart}, and Rankformer \cite{rankformer}, as LambdaMART is still considered the SOTA pairwise ranking method and Rankformer is the SOTA deep learning-based listwise ranking method. 
The benchmark models chosen include the best pure pairwise and pure listwise methods, which is sufficient to demonstrate the effectiveness of our approach.

In practice, we replace the original prediction layer, which predicts the ranking scores for each entity, with different benchmark models to evaluate the performance achieved by this modification.

\subsection{Datasets}

\paragraph{Stock Ranking}
The task of stock ranking focuses on accurately predicting and ranking stocks according to their anticipated future returns ratio, which aids investors in selecting stocks for investment purposes~\citep{temporal-stock-prediction, zhang2022constructing}. 
We use the dataset\footnote{\url{https://github.com/fulifeng/Temporal_Relational_Stock_Ranking/}} introduced by \citeauthor{temporal-stock-prediction} \cite{temporal-stock-prediction}, which includes historical trading data from 2013 to 2017 for two significant markets, NASDAQ and NYSE, containing 1026 and 1737 stocks, respectively.
We maintain a consistent setting for splitting the data into training, validation, and testing sets over a 3-year, 1-year, and 1-year period. 

\paragraph{Information Retrieval}
Information retrieval is another critical area where ranking models are extensively applied.
The task of information retrieval focuses on accurately ranking documents based on their relevance to a given query, which is crucial for search engines.
LambdaMART and Rankformer are both originally proposed for information retrieval task \cite{lambdamart,rankformer}.
We use the MQ2008-list\footnote{\url{https://www.microsoft.com/en-us/research/project/letor-learning-rank-information-retrieval/letor-4-0/}} dataset from Microsoft LETOR4.0 \cite{letor4}, a popular benchmark dataset for LETOR algorithms. 
We acknowledge the existence of more recent information retrieval datasets, but they only provide relevance-level labels without ordinal ranking labels, making them unsuitable for our model evaluation.

\paragraph{Historical Events Ordering}
In the final task, we include a textual dataset called ``On This Day 2'' (OTD2)\footnote{\url{https://github.com/ltorroba/machine-reading-historical-events}} for chronologically ordering historical events. 
Originally, \citeauthor{honovich-etal-2020-machine} constructed this dataset for a regression task to predict the year of occurrence \cite{honovich-etal-2020-machine}. 
However, it can also be approached as a ranking task to predict the chronological order of events within a given group of events. 
The OTD2 dataset, sourced from the ``On This Day'' website, includes 71,484 events enriched with additional contextual information. 
Scraped in April 2020, it incorporates recent events and corrections by the site. 
Preprocessing excluded events before 1 CE and future projections such as ``31st predicted perihelion passage of Halley's Comet'' in 2061 CE. 
We follow the 80\%/10\%/10\% train/validation/test split as provided.

\subsection{Backbones}

\paragraph{Stock Ranking}
We use the method by \citeauthor{temporal-stock-prediction} \cite{temporal-stock-prediction} as the backbone model, replacing the prediction layer with different benchmark ranking models.
Although other potential methods such as STHAN-SR~\cite{sawhney2021stock}, ALSP-TF~\cite{ijcai2022p551}, and CI-STHPAN~\cite{CI-STHPAN} have been proposed for stock selection, we do not include them due to the lack of last-state embeddings needed to integrate TSPRank, unavailability of source code, or reproducibility issues.
Since our primary goal is to demonstrate TSPRank as a versatile ranking approach rather than achieving SOTA results on a single dataset, this selection of backbone is sufficient to validate our model's performance.

\paragraph{Information Retrieval}
Given that the features in the MQ2008-list are encoded using a combination of BM25 and TF-IDF methods, we treat BM25 and TF-IDF as the backbone models and use these encoded features as embeddings.

\paragraph{Historical Events Ordering}
Concatenating event titles and information can result in contexts spanning thousands of tokens. 
Therefore, we use the \textit{text-embedding-3-small}\footnote{\url{https://platform.openai.com/docs/guides/embeddings}} embeddings from OpenAI as the backbone model, which supports up to 8191 tokens. 
Our goal is not to achieve SOTA performance but to ensure a fair comparison. 
The small version (\textit{text-embedding-3-small}) with 1536 dimensions, compared to the large version's 3072 dimensions (\textit{text-embedding-3-large}), is more computationally efficient and sufficient for our purposes.

\subsection{Technical Setup}

\paragraph{Stock Ranking}
Considering that TSPRank specialises in small-scale ranking problems, we implement rankings within individual sectors, categorising stocks into groups based on their sectors. 
We exclude all sector groups containing three or fewer stocks and remove stocks with unidentified sector information.
Detailed descriptions of the sector-specific grouping are provided in Appendix~\ref{appendix:stock-sectors}.
For the benchmarking process, we use outputs from the relational embedding layer introduced by \citeauthor{temporal-stock-prediction} \cite{temporal-stock-prediction}, replacing the original prediction layer with the other benchmark models. 
This modification enables us to assess the extent to which transitioning from a point-wise (original model) to pairwise, listwise, and our hybrid approaches can enhance performance. 
All benchmark models and TSPRank with local learning are trained for 100 epochs for each sector, as 100 epochs are sufficient for the benchmark models to converge. We find that TSPRank with global learning generally requires more epochs to converge; therefore, TSPRank-global is trained for an additional 50 epochs to ensure convergence. Other hyperparameters are specified in Appendix~\ref{appendix:hyperparameters}.
For LambdaMART, we use 10,000 trees since previous experiments indicated that Rankformer failed to outperform LambdaMART with this configuration \cite{rankformer}. 
We choose the widely-used XGBoost\footnote{\url{https://xgboost.readthedocs.io/en/latest/tutorials/learning_to_rank.html}} \cite{xgboost} implementation of LambdaMART, especially due to the performance enhancements in its version 2.0. % released in September 2023.

\paragraph{Information Retrieval}
In the MQ2008-list dataset, a query might contain over 1,000 documents, resulting in a graph that is too large for the TSP solver to handle efficiently. 
Therefore, we focus on a reranking or post-reranking stage. 
Specifically, we extract the top 10 and top 30 documents for each query to conduct two separate experiments. 
This approach allows us to manage the computational complexity while still evaluating the effectiveness of TSPRank in a realistic ranking scenario.
% For LambdaMART and Rankformer models, no additional processing is required. 
% However, for TSPRank, we preprocess the relevance labels by converting them to rankings through inversion, ensuring that lower label values correspond to higher relevance in the ranking. 
% This conversion aligns with TSPRank's use of the ascending order permutation as the predictive output. 
% When calculating the evaluation metrics, we then invert the permutation obtained from TSPRank back to the original relevance labels. 
% This ensures consistency and accuracy in the performance assessment of the model.
We use similar hyperparameters as in the stock ranking task but increase the number of transformer layers in both Rankformer and TSPRank to four, as we do not use any pretrained embeddings here.
To ensure fairness, we also follow the 5-fold cross-validation setup as provided by Microsoft.

\paragraph{Historical Events Ordering}
We randomly allocate the events into ranking groups and train the ranking models to predict the ground truth order of occurrences within these groups. 
We test with group sizes of 10, 30 and 50 for fair comparisons. 
For each group size, we generate the group allocations using five different random seeds and report the average performance to mitigate the effects of random group allocation.
Regarding the models, we observed overfitting with Rankformer and TSPRank when including the transformer block, even with a 0.5 dropout rate. 
This is likely due to the strong representational power of the contextual embedding and the relatively small size of the OTD2 dataset compared to other NLP corpora. 
Consequently, we retained only the scoring network in Rankformer and the trainable bilinear model in TSPRank, integrating them directly with the \textit{text-embedding-3-small} embeddings. 
This setting not only prevents overfitting but also ensures similar model sizes.
Other hyperparameters are specified in Appendix~\ref{appendix:hyperparameters}.
We remove the weights in the weighted loss function for TSPRank-Local here since all events should be weighted equally without any additional requirements in this task.

\subsection{Evaluation}

\paragraph{Stock Ranking}
Given that our task encompasses both ranking and stock selection functionalities, we use both traditional ranking metrics and financial performance metrics in our evaluation. 
For ranking effectiveness, we use the Mean Average Precision at $K$ (MAP@K), which evaluates the precision of the top $K$ ranked stocks by averaging the precision scores at each relevant stock position. 
We also use Kendall's Tau, defined as: $\tau = \frac{(C - D)}{\frac{1}{2} n (n-1)}$, where \( C \) is the number of concordant pairs (pairs of entities that are in the same order in both predicted and ground truth rankings), \( D \) is the number of discordant pairs (pairs of entities that are in different orders in the rankings), and \( n \) is the total number of entities being ranked.
Kendall's Tau is a correlation coefficient that measures the ordinal association between two ranked lists. This is suitable in our case, where we need to assess the similarity between the predicted rankings and the ground-truth rankings.
Furthermore, to assess the financial impact of the rankings, we simulate trading activities by predicting the rankings of all individual stocks in every sector for the next trading day $t+1$ on trading day $t$. 
This simulation is consistent with the one presented by \citeauthor{temporal-stock-prediction} \cite{temporal-stock-prediction}.
We then rank these stocks, hold the top $k$ stocks $S_k$, and sell them on trading day $t+1$. 
The price of stock $i$ on day $t$ is denoted as $p_i^{(t)}$ per share. 
The cumulative Investment Return Ratio (IRR) and Sharpe Ratio (SR) are defined in Equation~\ref{eq:irr-and-sr}, where $R_p$ is the return of the portfolio, $R_f$ is the risk-free rate, and $\sigma$ is the standard deviation:

\begin{equation}
\label{eq:irr-and-sr}
    \text{IRR}^k = \sum_{i \in S^k} \frac{p_i^{(t)} - p_i^{(t-1)}}{p_i^{(t-1)}}, \text{  SR} = \frac{E(R_p-R_f)}{\sigma(R_p)}.
\end{equation}

\paragraph{Information Retrieval}
We evaluate the performance of our models using several well-established ranking metrics: Normalized Discounted Cumulative Gain (NDCG) at various cutoff levels (3, 5, 10), Mean Reciprocal Rank (MRR), and Kendall's tau. 
NDCG measures a document's usefulness based on its position in the result list, with higher positions receiving more weight. 
% The formula for NDCG at position $k$ is defined in Equation~\ref{eq:ndcg}, where DCG@K is the Discounted Cumulative Gain at position $K$ and IDCG@K is the Ideal DCG at position $K$. 
% Additionally, the formula of DCG@K is also given in Equation~\ref{eq:ndcg}, where $\text{rel}_i$ denotes the relevance score of the document at position $i$. Specifically, IDCG@K is the maximum possible DCG up to position $K$, serving as a normalisation factor.
On the other hand, MRR assesses the rank position of the first relevant document, calculated as the average of the reciprocal ranks of the first relevant document for each query. The formula for MRR is defined by $\text{MRR} = \frac{1}{|Q|} \sum_{i=1}^{|Q|} \frac{1}{\text{rank}_i}$, where $|Q|$ is the total number of queries and $\text{rank}_i$ is the rank position of the first relevant document for the $i$-th query.

% \begin{equation}
% \label{eq:ndcg}
%     \text{NDCG@K} = \frac{\text{DCG@K}}{\text{IDCG@K}},\  \text{DCG@K} = \sum_{i=1}^{K} \frac{2^{\text{rel}_i} - 1}{\log_2(i+1)}\text{.}
% \end{equation}

\paragraph{Historical Events Ordering}
We report the Kendall's Tau, MRR, Exact Match (EM) of ordering, and the Root Mean Square Error (RMSE) of the predicted and ground truth rankings.
EM is defined as the ratio of the number of events whose predicted order exactly matches the ground truth order to the total number of events, formulated as: $\text{EM} = \frac{\text{\# correctly ordered events}}{\text{\# events}}$. 
% In this task, we consider MRR a less important metric since all events should be weighted equally without additional requirements.

\section{Results}
\label{sec:results}

\begin{table*}[htbp]
\resizebox{\textwidth}{!}{%
\begin{tabular}{llrrrrrrrrrr}
\toprule
Market                  & Model          & $\tau$             & IRR@1           & SR@1            & MAP@1           & IRR@3           & SR@3            & MAP@3           & IRR@5           & SR@5            & MAP@5           \\
\midrule
\multirow{5}{*}{NASDAQ} & \citeauthor{temporal-stock-prediction} + MLP (Original)      & 0.0093          & 0.1947          & 0.5341          & \textbf{0.1690} & 0.2366          & 0.9881          & 0.3253          & 0.1892          & 0.9682          & 0.5871          \\
                        & \citeauthor{temporal-stock-prediction} + LambdaMART    & 0.0071    & 0.0310    & -0.0873   & 0.1539    & 0.0340    & 0.0445    & 0.3144    & 0.0505    & 0.2678    & 0.5858   \\
                        & \citeauthor{temporal-stock-prediction} + Rankformer   & 0.0110        &  0.2257        &  0.5464     & 0.1620      &  0.2857       &    1.1245     &  0.3216       &  0.2309       &  1.0943       &  0.5860 \\
                        & \citeauthor{temporal-stock-prediction} + TSPRank-Local  & 0.0291          & 0.5353          & 1.2858          & 0.1658          & 0.4416          & 1.7401          & 0.3297          & 0.2537          & 1.2623         & 0.5932          \\
                        & \citeauthor{temporal-stock-prediction} + TSPRank-Global & \textbf{0.0447} & \textbf{0.7849} & \textbf{1.7471} & 0.1633          & \textbf{0.5224} & \textbf{2.0359} & \textbf{0.3364} & \textbf{0.2937} & \textbf{1.4331} & \textbf{0.5999} \\ \midrule
\multirow{5}{*}{NYSE}   & \citeauthor{temporal-stock-prediction} + MLP (Original)      & 0.0162          & 0.4170          & 1.0755          & \textbf{0.1791}          & 0.2574          & 1.2367          & \textbf{0.2841}          & 0.2257          & 1.3186          & 0.4649          \\
                        & \citeauthor{temporal-stock-prediction} + LambdaMART    & 0.0054    & 0.1005    & 0.1367    & 0.1307    & 0.0732    & 0.4192    & 0.2592    & 0.1063    & 0.6882    & 0.4574    \\
                        & \citeauthor{temporal-stock-prediction} + Rankformer    & 0.0181    & 0.2924    & 0.9113    & 0.1535    & 0.2701    & 1.2890    & 0.2758    & 0.2515    & 1.4200    & 0.4651 \\
                        & \citeauthor{temporal-stock-prediction} + TSPRank-Local  & 0.0313       & \textbf{0.5012} & \textbf{1.5710} & 0.1424 & \textbf{0.3974} & 1.9735 & 0.2756          & 0.2788          & 1.6662          & 0.4680          \\
                        & \citeauthor{temporal-stock-prediction} + TSPRank-Global & \textbf{0.0422} & 0.4787          & 1.4552          & 0.1392          & 0.3889          & \textbf{1.9976}          & 0.2756 & \textbf{0.2816} & \textbf{1.7350} & \textbf{0.4732} \\
\bottomrule
\end{tabular}%
}
\caption{Performance comparison of \citeauthor{temporal-stock-prediction}, LambdaMART, Rankformer, and TSPRank on the NASDAQ and NYSE stock ranking dataset, averaged across all filtered sectors.}
\label{tab:stock-results}
\vspace{-0.4cm}
\end{table*}

\begin{table*}[htbp]
\resizebox{\textwidth}{!}{%
\begin{tabular}{ll|rrrrr|rrrrr}
\toprule
        &   & \multicolumn{5}{c|}{Top 10}                  & \multicolumn{5}{c}{Top 30}                                                              \\ \midrule
Model & Type     & NDCG@3 & NDCG@5 & NDCG@10 & MRR    & $\tau$    & NDCG@3          & NDCG@5          & NDCG@10         & MRR             & $\tau$             \\ \midrule
LambdaMART & Pairwise & 0.6833 & 0.7222 & 0.8707  & 0.4259 & 0.1474 & 0.7340          & 0.7298          & 0.7403          & 0.3617          & 0.2372          \\
Rankformer & Listwise & 0.7220 & 0.7565 & 0.8865  & 0.4661 & \textbf{0.2317} & 0.7486          & 0.7470          & 0.7596 & 0.3732          & \textbf{0.2834} \\
TSPRank-Local & Pairwise-Listwise    &  0.6858   &  0.7213  &  0.8719  &  0.4266   &  0.1544  &  0.7189  &  0.7240   &   0.7362    &   0.3206    &   0.2054 \\
TSPRank-Global & Pairwise-Listwise     &  \textbf{0.7281}    &  \textbf{0.7585}  &  \textbf{0.8884}  &  \textbf{0.4861}  & 0.2212  & \textbf{0.7582} & \textbf{0.7558} & \textbf{0.7631}      & \textbf{0.3895} & 0.2647    \\ \bottomrule
\end{tabular}%
}
\caption{Evaluation of LambdaMART, Rankformer, and TSPRank on MQ2008-list information retrieval dataset for top 10 and top 30 documents.}
\label{tab:mq2008-results}
\vspace{-0.4cm}
\end{table*}

\begin{table*}[htbp]
\centering
\resizebox{\textwidth}{!}{%
\begin{tabular}{l|l|rrrr|rrrr|rrrr}
\toprule
\multicolumn{2}{c|}{Group Size} & \multicolumn{4}{c|}{10} & \multicolumn{4}{c|}{30} & \multicolumn{4}{c}{50} \\ \midrule
Model & Type & $\tau$ $\uparrow$ & EM $\uparrow$ & MRR $\uparrow$ & RMSE $\downarrow$ & $\tau$ $\uparrow$ & EM $\uparrow$ & MRR $\uparrow$ & RMSE $\downarrow$ & $\tau$ $\uparrow$ & EM $\uparrow$ & MRR $\uparrow$ & RMSE $\downarrow$ \\ \midrule
\textit{te-3-small} + LambdaMART & Pairwise  & 0.6297 & 0.3008 & 0.7554  & \textbf{1.993} & 0.5929 & 0.1064 & 0.6122 & 5.969 & 0.6000 & 0.0639 & 0.5596 & 9.618 \\
\textit{te-3-small} + Rankformer & Listwise  & 0.6190 & 0.2899 & 0.7361 & 1.998 & 0.5859 & 0.0921 & 0.4911 & 5.973 & 0.5724 & 0.0527 & 0.3526 & 10.069
\\
\textit{te-3-small} + TSPRank-Local & Pairwise-Listwise  & 0.5658 & 0.2856 & 0.7679 & 2.296 & 0.5095 & 0.0873 & 0.5739 & 6.930 & 0.4713 & 0.0460 & 0.3949 & 12.084 \\
\textit{te-3-small} + TSPRank-Global & Pairwise-Listwise  & \textbf{0.6301} & \textbf{0.3350} & \textbf{0.7936} & 2.057 & \textbf{0.6302} & \textbf{0.1384} & \textbf{0.7300} & \textbf{5.770} & \textbf{0.6207} & \textbf{0.0871} & \textbf{0.6618} & \textbf{9.602} \\ \bottomrule
\end{tabular}%
}
\caption{Evaluation of LambdaMART, Rankformer, and TSPRank on OTD2 dataset for historical events ordering for group sizes of 10, 30, and 50. ``\textit{te-3-small}'' stands for ``\textit{text-embedding-3-small}''.}
\label{tab:historical-events-results}
\vspace{-0.5cm}
\end{table*}

The aggregated results for all benchmark models on the stock ranking dataset are presented in Table~\ref{tab:stock-results}.
These results represent the average performance across 50 sectors in the NASDAQ market and 70 sectors in the NYSE market, excluding sectors with three or fewer stocks as outlined in the experimental setup. 
Detailed results for each individual sector are provided in Appendix~\ref{appendix:stock-sectors}.
The results on the MQ2008-list retrieval dataset are shown in Table~\ref{tab:mq2008-results}.
A 5-fold cross-validation was performed, with detailed results for each fold provided in Appendix~\ref{appendix:mq2008-complete}.
Lastly, the results on OTD2 for historical events ordering are presented in Table~\ref{tab:historical-events-results}, with complete results for different random group allocations provided in Appendix~\ref{appendix:events-complete}.

The results highlight the comparisons among three types of models: pairwise (LambdaMART), listwise (Rankformer), and our pairwise-listwise method (TSPRank). 
\\[4pt]
\textbf{Better Performance of Pairwise-Listwise Method Across Diverse Tasks (RQ1).}
Our pairwise-listwise method, TSPRank-Global, demonstrates outstanding robustness and superior performance across diverse datasets and domains. 
For instance, on the NASDAQ stock ranking dataset (Table~\ref{tab:stock-results}), TSPRank-Global achieves a Kendall's Tau of 0.0447, significantly surpassing both LambdaMART (0.0071) and Rankformer (0.0110). 
Similarly, on the NYSE dataset (Table~\ref{tab:stock-results}), TSPRank-Global attains a Kendall's Tau of 0.0422, outperforming LambdaMART (0.0054) and Rankformer (0.0181). 
Both TSPRank-Local and TSPRank-Global also consistently achieve higher IRR and SR, yielding improved financial performance.
The model's robustness is further evidenced by its consistent top performance across other datasets. 
For example, in the MQ2008-list (Table~\ref{tab:mq2008-results}), TSPRank-Global leads with NDCG@10 scores of 0.8884 for the top 10 documents and 0.7631 for the top 30 documents, indicating superior ranking accuracy, despite a minor difference in Kendall's Tau compared to Rankformer (around 0.01 to 0.02). 
Additionally, TSPRank-Global achieves top performance in the historical events ordering task (Table~\ref{tab:historical-events-results}), regardless of the group size (10, 30, and 50).
\\[4pt]
\textbf{TSPRank Can Be Extra Component Upon Embeddings For Boosting The Performance.}
In the stock ranking and ordering tasks for historical events, we directly deployed TSPRank on the stock embeddings and text embeddings. 
As demonstrated, TSPRank effectively serves as an additional component on embeddings, significantly enhancing ranking performance. 
This capability allows TSPRank to have broader applications across various domains.
\\[4pt]
\textbf{Stronger Robustness Towards Number of entities.} 
TSPRank-Global also shows greater robustness concerning the number of entities compared to the pure pairwise LambdaMART and listwise Rankformer. 
As evidenced by the MQ2008 (Table~\ref{tab:mq2008-results}) results, the gap between LambdaMART and Rankformer narrows as the number of documents increases from 10 to 30, with the NDCG@3 score difference between these two methods reducing from -0.0387 to -0.0146. 
Similarly, in historical events ordering (Table~\ref{tab:historical-events-results}), LambdaMART and Rankformer's performance fluctuates more significantly with group size changes. 
This is evident from the varying differences in their Kendall's Tau and MRR compared to those of TSPRank-Global.
Conversely, TSPRank-Global consistently performs well regardless of group size, maintaining top performance across most metrics.
\\[4pt]
\textbf{Global Learning Outperforms Local Learning (RQ2). }
As evidenced by the performance of TSPRank-Global compared to TSPRank-Local on MQ2008 and OTD2. 
For example, on the OTD2 dataset (Table~\ref{tab:historical-events-results}), TSPRank-Global achieves a Kendall's Tau of 0.6301, 0.6302, 0.6207 for the three different group sizes, compared to TSPRank-Local’s 0.5658, 0.5095, and 0.4713.
These results suggest that the end-to-end global optimisation approach of TSPRank-Global aligns more effectively with overall optimisation goals, whereas the non-end-to-end method may not perfectly align with global optimisation by the TSP solver. 
\\[4pt]
\textbf{LambdaMART Does Not Always Outperform Deep Learning Based Ranking Algorithm (RQ3). } 
Although \citeauthor{google-neural-outperform-gbdt} suggests that GBDT models, such as LambdaMART, generally outperform deep learning models on standard information retrieval datasets \cite{google-neural-outperform-gbdt}, our results indicate that this is not always the case in other contexts.
As demonstrated in Table~\ref{tab:stock-results}, within the domain-specific scenario of stock ranking, LambdaMART shows markedly poorer performance. 
Specifically, LambdaMART achieves a Kendall's Tau of 0.0071 on NASDAQ and 0.0054 on NYSE, trailing behind Rankformer (0.0110 and 0.0181) and TSPRank-Global (0.0447 and 0.0422).
Table~\ref{tab:mq2008-results} further validates this observation.
In both the top 10 and top 30 document settings, LambdaMART consistently underperforms Rankformer and TSPRank-Global when adapting to ordinal ranking labels across all metrics. 
These results underscore that while LambdaMART performs well with binary relevance labels or relevance levels labels, it is not always the optimal choice for recovering complete rankings.

Overall, TSPRank's success suggests that combining pairwise and listwise approaches helps the model capture the relative ordering of entities and the overall ranking structure. 
This dual focus likely contributes to its superior performance across various datasets and metrics, making it versatile and powerful.

\section{Visualisation Analysis}
\label{sec:visualisation-analysis}

As mentioned before, our pairwise-listwise TSPRank addresses both the lack of listwise optimisation in pairwise LambdaMART and the robustness issues in listwise Rankformer, which arise from attempting to predict the complete rankings directly.

\begin{figure}[htbp]
    \centering
    \includegraphics[width=\columnwidth]{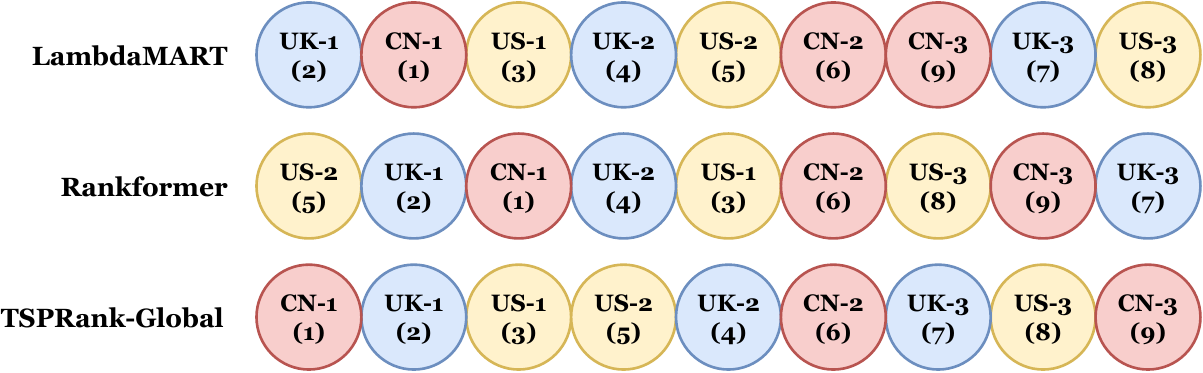}
    \caption{Visualisation of predictions by LambdaMART, Rankformer, and TSPRank-Global on the constructed group. Numbers in parentheses indicate the true ranking.}
    \label{fig:prediction-visualisation}
    \vspace{-0.4cm}
\end{figure}

To empirically validate this and answer \textbf{RQ4} (the advantages of hybrid pairwise-listwise TSPRank over other methods), we conducted another experiment on the OTD2 dataset, chosen because of the better interpretability provided by text.
We arbitrarily sample three events each from the US, UK, and China, ensuring that the events within each country occurred at least 50 years apart (detailed in Table~\ref{tab:manually-events-data}). 
We then use the pretrained LambdaMART, Rankformer, and TSPRank-Global from the experiments described in Section~\ref{sec:exp-setup} to predict the rankings for this constructed group.
We visualise the predictions given by the three rankers in Figure~\ref{fig:prediction-visualisation}.
Both the pairwise LambdaMART and pairwise-listwise TSPRank-Global successfully recover the intra-country event order, correctly identifying the relative order within the same colour group. 
However, the listwise Rankformer incorrectly predicts that ``US-2'' occurs earlier than ``US-1''.

Although LambdaMART successfully recovers the intra-country event order, it makes mistakes in the overall ranking, particularly failing to recover the correct order of [``US-1'', ``UK-1'', ``CN-1''] and [``US-3'', ``UK-3'', ``CN-3''], validating its weaker ability for listwise ranking optimisation. 
In contrast, TSPRank accurately recovers most orders, with only a minor error between ``US-2'' and ``UK-2'', which only have a 30-year gap (1934 and 1904).

\begin{figure}[htbp]
    \centering
    \includegraphics[width=\columnwidth]{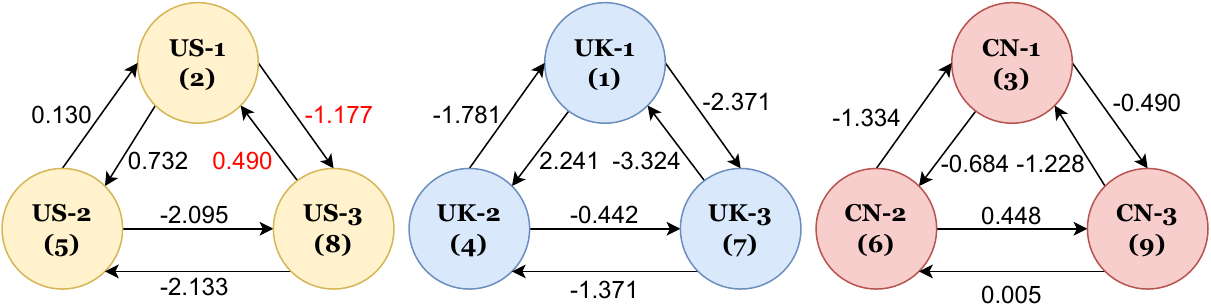}
    \caption{Illustration of the intra-country pairwise comparison graph. Edges between pairs of events from different countries are omitted for clarity. Scores highlighted in red indicate errors in the pairwise prediction for TSPRank-Global.}
    \label{fig:tsprank-visualisation}
    \vspace{-0.5cm}
\end{figure}

We also visualise the intra-country connections using the adjacency matrix predicted by TSPRank's bilinear model in Figure~\ref{fig:tsprank-visualisation}, omitting cross-country connections for clarity. 
As highlighted in red, the direct comparison between ``US-1'' and ``US-3'' is incorrect since $s(\text{US-1}, \text{US-3})$ should be higher than $s(\text{US-3}, \text{US-1})$, but the model fails in this case. 
However, as shown in Figure~\ref{fig:prediction-visualisation}, TSPRank still correctly places ``US-1'' at position 3 and ``US-3'' at position 8. 
This indicates that the model benefits from the cross-country connections and the listwise optimisation provided by the TSP solver. 
This makes the model more tolerant to errors or uncertainties in pairwise comparisons, as the listwise optimisation ensures that the highest-scoring permutation is selected.
We provide additional examples in Appendix~\ref{appendix:additional-examples} to further validate this observation.

\begin{table}[htbp]
    \centering
    \resizebox{\columnwidth}{!}{%
    \begin{tabular}{p{0.8\columnwidth}ccc}
    \toprule
    Event Title & Year & Rank & Label \\
    \midrule
    1st US store to install electric lights,   Philadelphia & 1878 & 3 & \yellow{US-1} \\ \midrule
    1st sitting US President to visit South America, FDR in Colombia & 1934 & 5 & \yellow{US-2} \\ \midrule
    75th US Masters Tournament, Augusta National GC: Charl Schwartzel of South Africa birdies the final 4 holes to win his first major title, 2 strokes ahead of Australian pair Adam Scott and Jason Day & 2011 & 8 & \yellow{US-3} \\ \midrule
    Charles Watson-Wentworth, 2nd Marquess of Rockingham, becomes Prime Minister of Great Britain & 1782 & 2 & \blue{UK-1} \\ \midrule
    1st main line electric train in UK (Liverpool to Southport) & 1904 & 4 & \blue{UK-2} \\ \midrule
    UK Terrorism Act 2006 becomes law & 2006 & 7 & \blue{UK-3} \\ \midrule
    A Mongolian victory at the naval Battle of Yamen ends the Song Dynasty in China & 1279 & 1 & \red{CN-1} \\ \midrule
    US Senate rejects China People's Republic membership to UN & 1953 & 6 & \red{CN-2} \\ \midrule
    China's Hubei province, the original center of the coronavirus COVID-19 outbreak eases restrictions on travel after a nearly two-month lockdown & 2020 & 9 & \red{CN-3} \\ \bottomrule
    \end{tabular}%
    }
    \caption{Event titles in the constructed group. Labels indicate the order of occurrence within each country, e.g., ``US-1'' denotes the earliest event in the US within the group.}
    \label{tab:manually-events-data}
    \vspace{-0.8cm}
\end{table}
% TODO color background for country label

% \begin{table}[htbp]
% \centering
% \resizebox{\columnwidth}{!}{%
% \begin{tabular}{llrrrr}
% \toprule
% Model & Type & Tau $\uparrow$ & EM $\uparrow$ & MRR $\uparrow$ & RMSE $\downarrow$ \\ \midrule
% LambdaMART & Pairwise & 0.833 & 0.444 & 0.500 & 0.943 \\
% Rankformer & ListWise & 0.556 & 0.333 & 0.333 & 1.826 \\
% TSPRank-Global & Pairwise-Listwise & \textbf{0.944} & \textbf{0.778} & \textbf{1.000} & \textbf{0.471} \\ \bottomrule
% \end{tabular}%
% }
% \caption{Performance on the manually constructed group.}
% \label{tab:manually-group-results}
% \end{table}

\section{Inference Latency}

Incorporating a combinatorial optimisation solver is known to cause additional computational overhead, which can impact the inference latency. 
Therefore, we compare the average inference time per ranking group for TSPRank and Rankformer on the OTD2 historical events ordering dataset. 
The experimental setup and detailed results are provided in Appendix~\ref{appendix:latency}. 

We observe that the inference time for TSPRank increases exponentially with the number of ranking entities, primarily due to the discrete TSP solver. 
When the group size reaches 100, the solver consumes approximately 99.8\% of the inference time, while the encoding time remains similar to that of Rankformer.
Therefore, TSPRank is more suitable for small-scale ranking problems with fewer than 30 entities, where its inference time is comparable to Rankformer.
It is also advantageous in scenarios where TSPRank’s performance improvements outweigh the increased latency. 
Solvers like Gurobi are efficient for small TSP graphs. 
However, exact solutions for larger graphs require specialised techniques, such as the Concorde TSP Solver\footnote{\url{https://www.math.uwaterloo.ca/tsp/concorde/index.html}}, which uses cutting-plane algorithms and a branch-and-bound approach to solve graphs with up to 85,900 nodes \cite{applegate2006traveling}. 
Beyond exact solvers, heuristic algorithms like the Christofides–Serdyukov algorithm \cite{christofides2022worst} and the Lin–Kernighan–Helsgaun heuristic \cite{lin1973effective,helsgaun2000effective} are also used.

Moreover, recent advances in neural networks have enabled the use of deep learning to approximate the TSP reward function, using reinforcement learning to solve the problem \cite{bello2016neural,khalil2017learning,kool2018attention}.
While we do not delve deeply into the various TSP solutions in this paper, we demonstrate that inference latency stems from the TSP solver and discuss potential enhancements for reducing inference latency and tackling large-scale ranking problems.

\section{Conclusion}

This study introduces TSPRank, a novel pairwise-listwise approach for position-based ranking tasks, by modelling them as the Travelling Salesman Problem (TSP).
We present two learning methods for TSPRank, integrating listwise optimization into pairwise comparisons to address the limitations of traditional pure pairwise and listwise ranking models. 
TSPRank's application to diverse backbone models and datasets, including stock ranking, information retrieval, and historical events ordering, demonstrates its superior performance and robustness. Key findings include TSPRank's ability to outperform existing models such as LambdaMART and Rankformer across multiple datasets and various metrics. 
Furthermore, TSPRank's capability to function as an additional component on embeddings suggests its versatility and potential for broader applications across different domains.

Our empirical analysis reveals that TSPRank’s superior performance is due to the discrete TSP solver’s ability to more effectively use listwise information.
It also enhances TSPRank’s tolerance to uncertainties in pairwise comparisons by ensuring the selection of the highest-scoring permutation. 
Future work could explore the use of alternative TSP solvers and further enhancements to the Gurobi solver to reduce inference latency and improve scalability.

%%
%% The next two lines define the bibliography style to be used, and
%% the bibliography file.
\bibliographystyle{ACM-Reference-Format}
\bibliography{sample-base}

\newpage
%%
%% If your work has an appendix, this is the place to put it.
\appendix

\section{TSP Optimisation Problem}
\label{appendix:tsp-constraints}

Ranking entity $\mathbf{e}_j$ immediately after entity $\mathbf{e}_i$ is analogous to travelling from city $i$ to city $j$.
Let $N$ be the total number of entities to be ranked, $s_{ij}$ be the abbreviation of $s(\mathbf{e}_i, \mathbf{e}_j)$ and $z_i$ variables to represent the number of entities ranked before entity $i$. 

\begin{align}
    \max\limits_{x_{ij}} & \quad \sum\limits_{i=1}^N \sum\limits_{j=1, j \neq i}^N s_{ij} x_{ij} \label{ilp-obj} \\
    s.t. & \quad \sum_{j=1, j \neq i}^N x_{ij} \leq 1  \quad \text{ for all } i \label{ilp-constr1} \\
    & \quad \sum_{i=1, i \neq j}^N x_{ij} \leq 1 \quad \text{ for all } j \label{ilp-constr2} \\
    & \quad \sum_{i=1}^N\sum_{j=1, j \neq i}^N x_{ij} = N-1 \label{ilp-constr3}\\
    & \quad z_i + 1 \leq z_j + N(1 - x_{ij}) \quad  i, j = 2, \ldots, N, \; i \neq j \label{ilp-constr4} \\
    &  \quad z_i \geq 0 \quad i = 2, \ldots, N\label{ilp-constr5}
\end{align}

Constraints (\ref{ilp-constr1}) and (\ref{ilp-constr2}) ensure that each entity has at most one predecessor and one successor in the ranking, expressed as $2N-2$ linear inequalities. Constraint (\ref{ilp-constr3}) ensures that the total number of pairwise comparisons is exactly $N-1$. These constraints ensure that the selected set of pairwise comparisons forms a valid ranking sequence. Constraints (\ref{ilp-constr4}) and (\ref{ilp-constr5}) introduce variables $z$ to eliminate multiple separate sequences and enforce that there is a single, complete ranking that includes all entities.

\section{Hyperparameters Setting}
\label{appendix:hyperparameters}

The hyperparameters are specified in Table~\ref{tab:hyperparameters}.
In the table, several abbreviations are used for conciseness and clarity:

\begin{itemizesquish}{0em}{0em}
    \item \textbf{lr}: Learning rate, which controls the step size during the optimization process.
    \item \textbf{tf layer}: Number of transformer layers, indicating the depth of the transformer model.
    \item \textbf{tf nheads}: Number of attention heads, which specifies the number of parallel attention mechanisms within the transformer layer.
    \item \textbf{tf dim\_ff}: Dimensionality of the feed-forward network within the transformer layer.
\end{itemizesquish}

\begin{table}[htbp]
\centering
\resizebox{\columnwidth}{!}{%
\begin{tabular}{l|l|l|l}
\toprule
\textbf{Dataset} & \textbf{Hyperparameter} & \textbf{Value} & \textbf{Model} \\ \hline
\multirow{5}{*}{All Tasks} & n\_estimators & $10^4$ & \multirow{4}{*}{LambdaMART} \\ \cline{2-3}
 & loss\_fn & rank:pairwise &  \\ \cline{2-3}
 & eval\_metric & auc &  \\ \cline{2-3}
 & early\_stopping & 50 &  \\ \cline{2-4}
 & loss & OrdinalLoss \cite{rankformer} & Rankformer \\ \hline
\multirow{6}{*}{Stocks} & lr & 1e-4 & \multirow{6}{*}{Rankformer, TSPRank} \\ \cline{2-3}
 & weight\_decay & 1e-5 &  \\ \cline{2-3}
 & tf layer & 1 &  \\ \cline{2-3}
 & tf nheads & 8 &  \\ \cline{2-3}
 & tf dim\_ff & 128 &  \\ \cline{2-3}
 & batch\_size & 128 &  \\ \hline
\multirow{6}{*}{MQ2008} & lr & 1e-4 & \multirow{6}{*}{Rankformer, TSPRank} \\ \cline{2-3}
 & weight\_decay & 1e-5 &  \\ \cline{2-3}
 & tf layer & 4 &  \\ \cline{2-3}
 & tf nheads & 8 &  \\ \cline{2-3}
 & tf dim\_ff & 128 &  \\ \cline{2-3}
 & batch size & 64 &  \\ \hline
\multirow{4}{*}{Historical Events} & lr & 1e-4 & \multirow{4}{*}{Rankformer, TSPRank} \\ \cline{2-3}
 & weight\_decay & 1e-5 &  \\ \cline{2-3}
 & tf dim\_ff & 0 &  \\ \cline{2-3}
 & batch size & 32 &  \\ \bottomrule
\end{tabular}%
}
\caption{Hyperparameters for LambdaMART model across different tasks: Stock Ranking, MQ2008, and Historical Events Ordering.}
\vspace{-0.5cm}
\label{tab:hyperparameters}
\end{table}

\section{Additional Visualisation Examples}
\label{appendix:additional-examples}

\begin{figure}[htbp]
    \centering
    \includegraphics[width=\columnwidth]{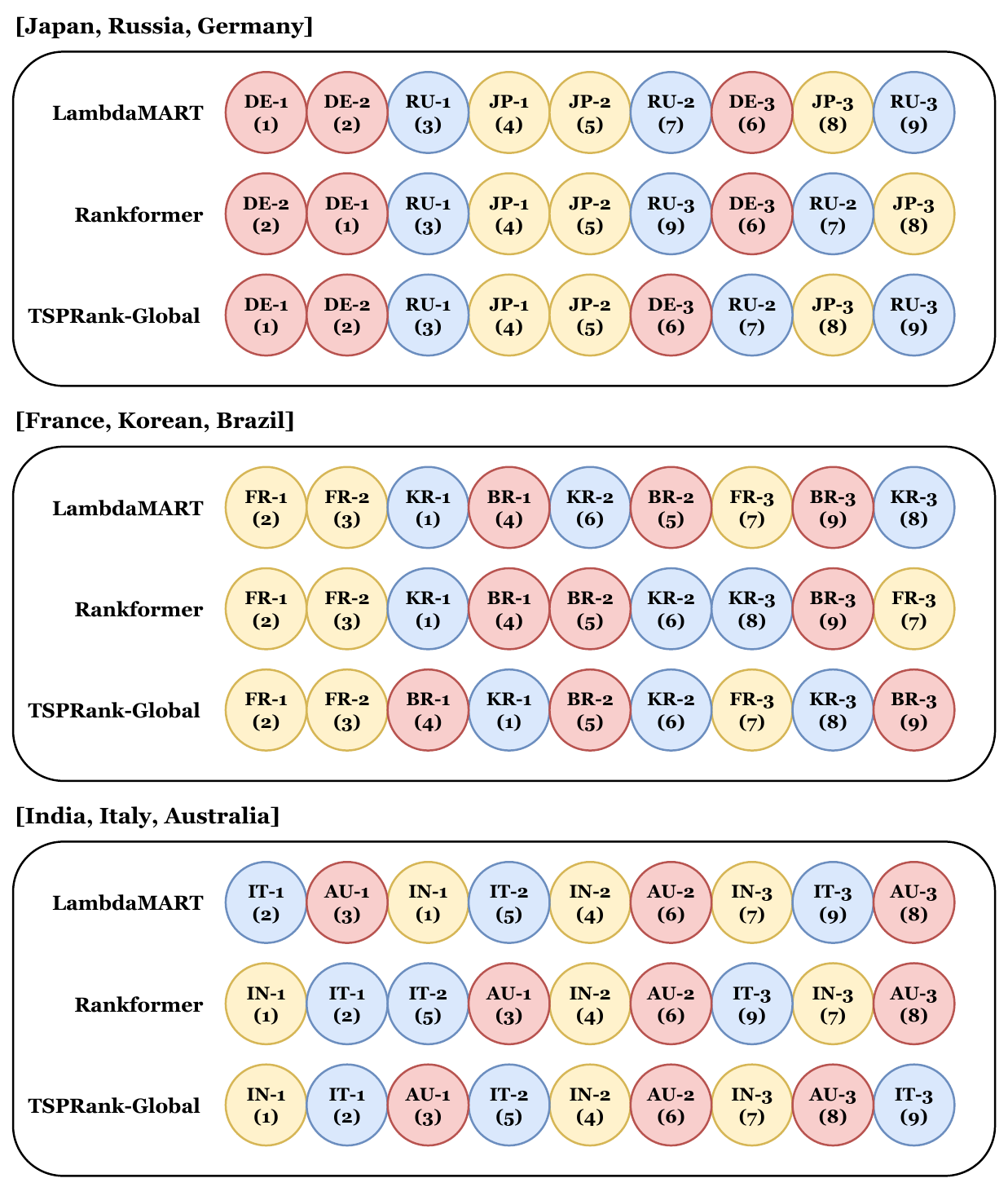}
    \caption{Visualisation of the model predictions on additional groups. Numbers in parentheses indicate the true ranking.}
    \label{fig:additional-visualisation}
    \vspace{-0.2cm}
\end{figure}

We further construct three more groups using events from different countries.
The event titles, labels, and true rankings are provided in Table~\ref{tab:addtional-groups}.
The visualisation of the models' predictions are shown in Figure~\ref{fig:additional-visualisation}.
We observe similar patterns to those discussed in Section~\ref{sec:visualisation-analysis}.

\begin{table*}[htbp]
\centering
\resizebox{\textwidth}{!}{%
\begin{tabular}{c|p{0.9\textwidth}ccc}
\toprule
Group & Event Title & Year & Rank & Label \\ \midrule
\multirow{9}{*}{1} & Japanese immigration to Brazil begins when 781 people arrive in Santos aboard the Kasato-Maru ship & 1908 & 4 & JP-1 \\ \cline{2-5} 
 & Passenger ship Lurline sends radio signal of sighting Japanese war fleet & 1941 & 5 & JP-2 \\ \cline{2-5} 
 & A reactor at the Fukushima Daiichi nuclear power plant melts and explodes and releases radioactivity into the atmosphere a day after Japan's earthquake. & 2011 & 8 & JP-3 \\ \cline{2-5} 
 & The State Bank of the Russian Empire is established. & 1860 & 3 & RU-1 \\ \cline{2-5} 
 & "Mirror", Russian film directed by Andrei Tarkovsky, starring Margarita Terekhova and Ignat Daniltsev, is released & 1975 & 7 & RU-2 \\ \cline{2-5} 
 & Russian city Moscow begins a city-wide lockdown after 4 hours notice due to coronavirus COVID-19 & 2020 & 9 & RU-3 \\ \cline{2-5} 
 & Albrecht II of Habsburg becomes king of Germany & 1438 & 1 & DE-1 \\ \cline{2-5} 
 & Magdeburg in Germany seized by forces of the Holy Roman Empire under earl Johann Tilly, most inhabitants massacred, one of the bloodiest incidents of the Thirty Years' War & 1631 & 2 & DE-2 \\ \cline{2-5} 
 & The cult classic "One Million Years B.C.", starring Raquel Welch, is released 1st in West Germany & 1966 & 6 & DE-3 \\ \midrule
\multirow{9}{*}{2} & French King Charles VIII occupies Florence & 1494 & 2 & FR-1 \\ \cline{2-5} 
 & 1st steamboat, Pyroscaphe, 1st run in France & 1783 & 3 & FR-2 \\ \cline{2-5} 
 & Francois Mitterrand becomes president of France & 1981 & 7 & FR-3 \\ \cline{2-5} 
 & The Hangul alphabet is published in Korea & 1446 & 1 & KR-1 \\ \cline{2-5} 
 & Korea is divided into North and South Korea along the 38th parallel & 1945 & 6 & KR-2 \\ \cline{2-5} 
 & North Korea blocks a South Korean supply delegation from the Kaesong joint industrial zone & 2013 & 8 & KR-3 \\ \cline{2-5} 
 & Bahia Independence Day: the end of Portuguese rule in Brazil, with the final defeat of the Portuguese crown loyalists in the province of Bahia & 1823 & 4 & BR-1 \\ \cline{2-5} 
 & Belo Horizonte, the first planned city of Brazil, founded & 1897 & 5 & BR-2 \\ \cline{2-5} 
 & Brazilian court blocks President Michel Temer from abolishing Renca, which would open parts of the Amazon to mining & 2017 & 9 & BR-3 \\ \midrule
\multirow{9}{*}{3} & Indian Mutiny against rule by the British East India Company begins with the revolt of the Sepoy soldiers in Meerut & 1857 & 1 & IN-1 \\ \cline{2-5} 
 & Gandhi supports the African People's Organisations resolution to declare the Prince of Wales day of arrival in South Africa a day of mourning, in protest against the South Africa Acts disenfranchisement of Indians, Coloureds and Africans & 1910 & 4 & IN-2 \\ \cline{2-5} 
 & Bomb attack on train in Assam India (27 soldiers killed) & 1995 & 7 & IN-3 \\ \cline{2-5} 
 & 1st Italian Parliament meets at Turin & 1860 & 2 & IT-1 \\ \cline{2-5} 
 & King Victor Emmanuel III of Italy abdicates and is succeeded by his son Umberto II who reigns for only 34 days before the monarchy is abolished & 1946 & 5 & IT-2 \\ \cline{2-5} 
 & Storms in Italy kill at least 11 with 75\% of Venice flooded and two tornadoes striking Terracina & 2018 & 9 & IT-3 \\ \cline{2-5} 
 & Edmund Barton is elected Prime Minister in Australia's first parliamentary election & 1901 & 3 & AU-1 \\ \cline{2-5} 
 & Australian Championships Women's Tennis: Beryl Penrose wins her only Australian singles title; beats Thelma Coyne Long 6-4, 6-3 & 1955 & 6 & AU-2 \\ \cline{2-5} 
 & Cricket World Cup, Melbourne (MCG): Australia defeats fellow host New Zealand by 7 wickets to win their 5th title; Player of Series: Mitchel Starc & 2015 & 8 & AU-3 \\ \bottomrule
\end{tabular}%
}
\caption{Three additionally constructed groups using events from different countries.}
\label{tab:addtional-groups}
\end{table*}

\section{Latency Analysis Details}
\label{appendix:latency}

We conduct an inference time analysis on a standalone device with an AMD Ryzen 7 5800X 8-Core CPU and an Nvidia RTX 4070 Super GPU. 
Using the standalone device instead of the one used for training is to avoid interference from other running jobs on the shared computing node.

\begin{figure}[htbp]
    \centering
    \includegraphics[width=\columnwidth]{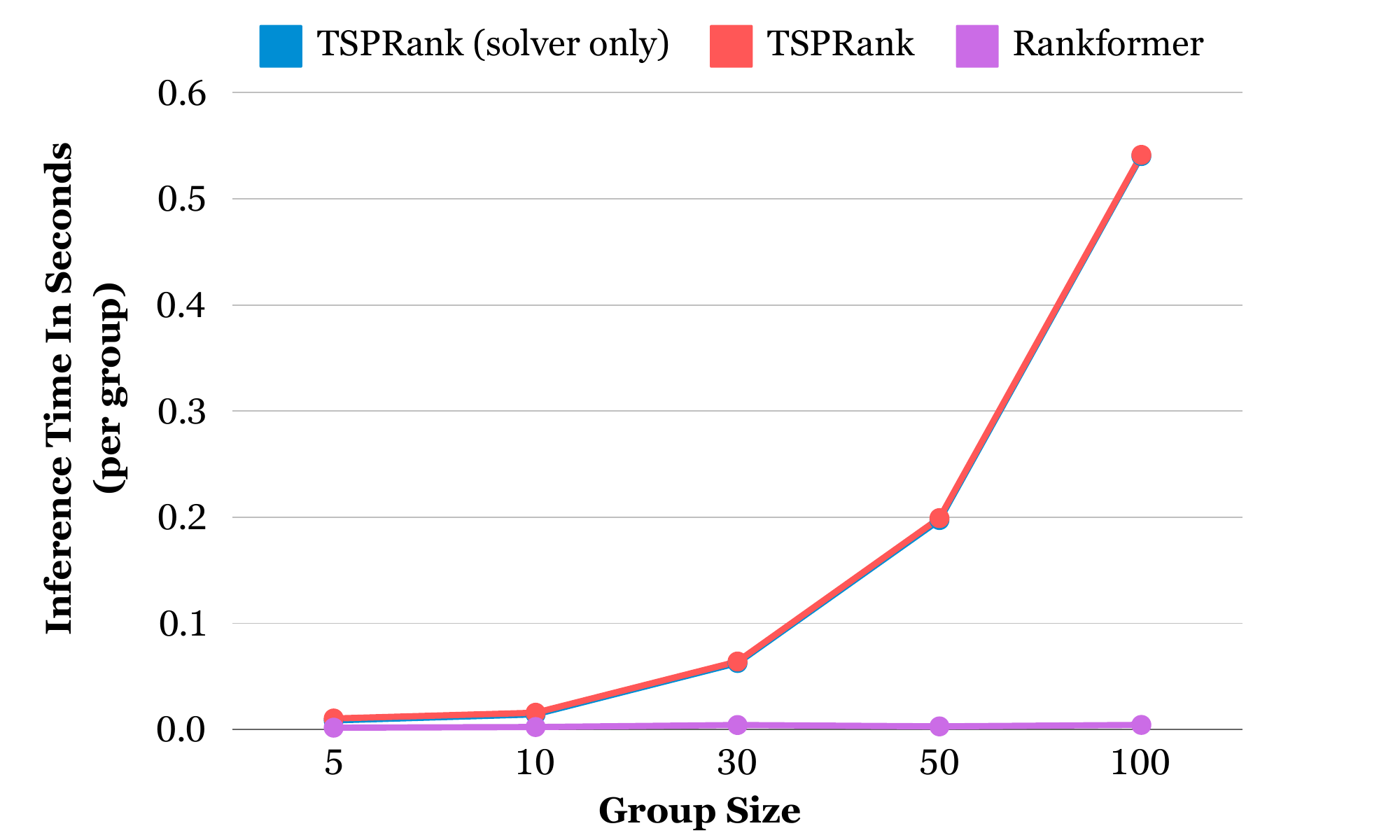}
    \caption{Mean inference time per ranking group for TSPRank and Rankformer on OTD2 across group sizes of 5, 10, 30, 50, and 100. The blue (TSPRank solver only) and red (TSPRank) lines overlap with minor differences.} 
    \label{fig:latency}
\end{figure}

Figure~\ref{fig:latency} illustrates the average inference time in seconds per ranking group for TSPRank and Rankformer on the OTD2 historical events ordering dataset, with an input dimensionality of 1536. 
We provide the inference times for various group sizes: 5, 10, 30, 50, and 100.

The comparison shows that the inference time for TSPRank increases exponentially with the number of ranking entities in a group. 
This increase is primarily due to the discrete TSP solver, as indicated by the overlapping blue and red lines. 
The blue line represents the time taken by the TSP solver, while the red line shows the total inference time for a ranking group. 
When the group size reaches 100, approximately 99.8\% of the inference time is consumed by the solver. 
The encoding time from the input to the adjacency matrix remains similar to that of Rankformer (purple).

% Thus, we consider TSPRank as a better solution for small-scale ranking problems with fewer than 30 ranking entities, since its inference time in this range is comparable to Rankformer. 
% However, we also see potential for improvement in handling larger-scale problems.
% Recent works have explored using deep learning techniques, such as convolutional neural networks with reinforcement learning, to solve TSP \cite{8659266}. 
% Another notable work, Tspformer \cite{YANG2023589}, employs the transformer model for TSP, achieving much faster inference speeds even for large-scale graphs with 500, 750, and 1000 nodes. 
% Tspformer reportedly solves TSP with 100 nodes in 0.035 seconds using a greedy search per graph, compared to 0.540 seconds for the Gurobi solver, without sacrificing performance. 
% It can also solve a TSP with 1000 nodes (ranking entities) in 0.146 seconds, which is even faster than the Gurobi solver for 50 nodes at 0.197 seconds.

% While we have not exhaustively tested different TSP solvers, opting instead for the well-known and commercially proven Gurobi solver for proof of concept, the latency issue can potentially be overcome using these alternative solvers. 
% We believe that further enhancements to Gurobi will also reduce the inference time.

\section{Complete Stocks Ranking Results For Sectors}
\label{appendix:stock-sectors}

Table~\ref{tab:baseline-results-nasdaq} to \ref{tab:global-training-results-nyse} presents the comprehensive results for each sector in NASDAQ and NYSE after filtering.

% ========================= baseline stock result ====================================
\begin{table*}[htbp]
\centering
\resizebox{\textwidth}{!}{%
\begin{tabular}{llrrrrrrrrrr}
\toprule
Sector                            & Stocks & $\tau$     & IRR@1   & SR@1    & MAP@1  & IRR@3   & SR@3    & MAP@3  & IRR@5   & SR@5    & MAP@5  \\ \midrule
Computer Manufacturing            & 5         & -0.0521 & -0.3767 & -1.2666 & 0.1857 & 0.1037  & 0.5894  & 0.4529 & 0.2167  & 1.3724  & 1.0000 \\
Transportation Services           & 5         & -0.0185 & -0.3402 & -1.1114 & 0.1941 & 0.2663  & 1.2239  & 0.4590 & 0.1263  & 0.7757  & 1.0000 \\
Medical/Dental Instruments        & 17        & 0.0227  & -0.1845 & -0.2096 & 0.1055 & 0.2604  & 1.2060  & 0.1519 & 0.3940  & 2.0538  & 0.1922 \\
Property-Casualty Insurers        & 19        & 0.0044  & -0.3242 & -0.5479 & 0.1055 & -0.1942 & -0.4679 & 0.2075 & -0.2643 & -1.0239 & 0.2131 \\
Medical Specialities              & 5         & -0.0101 & -0.0157 & 0.0948  & 0.1814 & 0.0702  & 0.4433  & 0.4902 & -0.0402 & -0.1500 & 1.0000 \\
Biotechnology: Biological Products (No Diagnostic Substances)  & 12 & 0.0288  & 0.5664  & 1.1922  & 0.1224 & 0.7839  & 2.1772  & 0.2253 & 0.4882  & 1.7150  & 0.3138 \\
Telecommunications Equipment                                   & 16 & -0.0046 & -0.6169 & -1.1101 & 0.1646 & -0.2057 & -0.4332 & 0.1718 & -0.1449 & -0.4478 & 0.2132 \\
\bottomrule
\end{tabular}%
}
\caption{Performance of baseline model proposed in \cite{temporal-stock-prediction} on NASDAQ}
\label{tab:baseline-results-nasdaq}
\end{table*}

\begin{table*}[htbp]
\centering
\resizebox{0.9\textwidth}{!}{%
\begin{tabular}{llrrrrrrrrrr}
\toprule
Sector                            & Stocks & $\tau$     & IRR@1   & SR@1    & MAP@1  & IRR@3   & SR@3    & MAP@3  & IRR@5   & SR@5    & MAP@5  \\ \midrule
Biotechnology: Laboratory Analytical Instruments & 5 & 0.0244 & 0.9165 & 2.6132 & 0.2447 & 0.5556 & 3.2469 & 0.4930 & 0.4553 & 3.1498 & 1.0000 \\
Diversified Commercial Services & 9 & 0.0315 & 1.9463 & 2.2132 & 0.2827 & 1.0483 & 2.8712 & 0.3033 & 0.7726 & 2.9765 & 0.3992 \\
Other Specialty Stores & 11 & -0.0083 & -0.2757 & -0.4046 & 0.1350 & 0.1972 & 0.7898 & 0.2220 & 0.2008 & 0.9486 & 0.3136 \\
Investment Managers & 18 & 0.0147 & 1.2026 & 3.1766 & 0.1224 & 0.7153 & 3.3275 & 0.1547 & 0.5711 & 3.2101 & 0.2090 \\
Electrical Products & 13 & 0.0021 & 0.2232 & 0.7774 & 0.1266 & 0.1747 & 0.9553 & 0.1699 & 0.2038 & 1.2445 & 0.2505 \\
Major Pharmaceuticals & 18 & 0.0083 & -0.1707 & -0.0414 & 0.1730 & -0.1365 & -0.3281 & 0.2218 & -0.0340 & -0.0488 & 0.2218 \\
Other Pharmaceuticals & 5 & 0.0420 & 0.7478 & 2.6057 & 0.2700 & 0.2195 & 1.2131 & 0.4916 & 0.1719 & 1.0601 & 1.0000 \\
\bottomrule
\end{tabular}%
}
\caption{Performance of baseline model proposed in \cite{temporal-stock-prediction} on NYSE}
\label{tab:baseline-results-nyse}
\end{table*}

% ======================= LambdaMART stock result ==============================
\begin{table*}[htbp]
\centering
\resizebox{\textwidth}{!}{%
\begin{tabular}{llrrrrrrrrrr}
\toprule
Sector & Stocks &  $\tau$ & IRR@1 & SR@1 & MAP@1 & IRR@3 & SR@3 & MAP@3 & IRR@5 & SR@5 & MAP@5 \\
\midrule
Computer Manufacturing & 5 & 0.0143 & -0.0477 & 0.0634 & 0.2658 & 0.1977 & 1.0821 & 0.4838 & 0.1648 & 1.0999 & 1.0000 \\
Transportation Services & 5 & 0.0303 & -0.3902 & -1.3876 & 0.2405 & -0.3105 & -1.6243 & 0.4913 & 0.0701 & 0.4840 & 1.0000 \\
Medical/Dental Instruments & 17 & 0.0164 & 1.2784 & 2.6419 & 0.0591 & 0.2390 & 1.0163 & 0.1275 & 0.0349 & 0.2789 & 0.1781 \\
Property-Casualty Insurers & 19 & 0.0438 & -0.4535 & -0.9007 & 0.1392 & -0.2711 & -1.1097 & 0.1636 & -0.2591 & -1.3919 & 0.1832 \\
Medical Specialities & 5 & 0.0286 & -0.3505 & -0.5935 & 0.2321 & -0.3086 & -1.5448 & 0.5000 & -0.1143 & -0.6432 & 1.0000 \\
Biotechnology: Biological Products (No Diagnostic Substances) & 12 & 0.0167 & -0.5780 & -0.9332 & 0.1224 & -0.3050 & -1.1125 & 0.1941 & -0.1622 & -0.6921 & 0.2982 \\
Telecommunications Equipment & 16 & -0.0026 & -0.7386 & -1.5655 & 0.1435 & -0.4117 & -1.4491 & 0.1535 & -0.3187 & -1.4648 & 0.1932 \\
\bottomrule
\end{tabular}%
}
\caption{Complete result table for LambdaMART on NASDAQ.}
\label{tab:lambdamart-training-results-nasdaq}
\end{table*}

\begin{table*}[htbp]
\centering
\resizebox{0.9\textwidth}{!}{%
\begin{tabular}{llrrrrrrrrrr}
\toprule
Sector & Stocks &  $\tau$ & IRR@1 & SR@1 & MAP@1 & IRR@3 & SR@3 & MAP@3 & IRR@5 & SR@5 & MAP@5 \\
\midrule
Biotechnology: Laboratory Analytical Instruments & 5 & 0.0000 & 0.6418 & 3.0829 & 0.1899 & 0.4841 & 3.4649 & 0.4970 & 0.4841 & 3.4492 & 1.0000 \\
Diversified Commercial Services & 9 & 0.0128 & 1.5345 & 2.2816 & 0.1983 & 0.6707 & 2.3281 & 0.2647 & 0.4743 & 2.2261 & 0.4146 \\
Other Specialty Stores & 11 & 0.0016 & -0.4049 & -0.7317 & 0.1435 & -0.2758 & -0.8344 & 0.2161 & -0.1138 & -0.3821 & 0.3255 \\
Investment Managers & 18 & 0.0280 & -0.0565 & -0.0769 & 0.0970 & 0.2047 & 1.1636 & 0.1484 & 0.3683 & 2.2241 & 0.1888 \\
Electrical Products & 13 & 0.0080 & -0.0764 & -0.2386 & 0.1055 & -0.0166 & -0.0226 & 0.1578 & 0.0685 & 0.5614 & 0.2450 \\
Major Pharmaceuticals & 18 & 0.0318 & 0.7983 & 1.4109 & 0.1308 & 0.1837 & 0.8006 & 0.1835 & 0.1154 & 0.6883 & 0.2122 \\
Other Pharmaceuticals & 5 & -0.0143 & -0.0489 & -0.0297 & 0.2025 & -0.0429 & -0.1446 & 0.4712 & 0.1293 & 0.8351 & 1.0000 \\
\bottomrule
\end{tabular}%
}
\caption{Complete result table for LambdaMART on NYSE.}
\label{tab:lambdamart-training-results-nyse}
\end{table*}

% ======================= Rankformer stock result ==============================
\begin{table*}[htbp]
\centering
\resizebox{\textwidth}{!}{%
\begin{tabular}{llrrrrrrrrrr}
\toprule
Sector & Stocks &  $\tau$ & IRR@1 & SR@1 & MAP@1 & IRR@3 & SR@3 & MAP@3 & IRR@5 & SR@5 & MAP@5 \\
\midrule
Computer Manufacturing & 5 & -0.0244 & -0.0458 & 0.0585 & 0.2110 & 0.2554 & 1.1537 & 0.4627 & 0.2167 & 1.3724 & 1.0000 \\
Transportation Services & 5 & -0.0193 & -0.3402 & -1.1114 & 0.1941 & 0.2787 & 1.2932 & 0.4557 & 0.1263 & 0.7757 & 1.0000 \\
Medical/Dental Instruments & 17 & 0.0492 & 0.1590 & 0.6044 & 0.1055 & 0.6454 & 2.2254 & 0.1592 & 0.6764 & 2.7925 & 0.2023 \\
Property-Casualty Insurers & 19 & 0.0136 & 0.1661 & 0.6221 & 0.0928 & -0.0236 & 0.0176 & 0.1306 & -0.1705 & -0.7351 & 0.1768 \\
Medical Specialities & 5 & 0.0336 & -0.0239 & 0.1681 & 0.2405 & 0.2459 & 1.1755 & 0.4768 & -0.0402 & -0.1500 & 1.0000 \\
Biotechnology: Biological Products (No Diagnostic Substances) & 12 & 0.0395 & 1.4993 & 2.0935 & 0.1519 & 1.4277 & 3.3088 & 0.2429 & 0.5337 & 1.8510 & 0.3092 \\
Telecommunications Equipment & 16 & -0.0179 & -0.7585 & -1.6430 & 0.1688 & -0.2981 & -0.7723 & 0.1779 & -0.1275 & -0.3238 & 0.2197 \\
\bottomrule
\end{tabular}%
}
\caption{Complete result table for Rankformer on NASDAQ.}
\label{tab:rankformer-training-results-nasdaq}
\end{table*}

\begin{table*}[htbp]
\centering
\resizebox{0.9\textwidth}{!}{%
\begin{tabular}{llrrrrrrrrrr}
\toprule
Sector & Stocks &  $\tau$ & IRR@1 & SR@1 & MAP@1 & IRR@3 & SR@3 & MAP@3 & IRR@5 & SR@5 & MAP@5 \\
\midrule
Biotechnology: Laboratory Analytical Instruments & 5 & -0.0025 & 0.6902 & 2.1537 & 0.1983 & 0.5843 & 3.4616 & 0.4698 & 0.4553 & 3.1498 & 1.0000 \\
Diversified Commercial Services & 9 & 0.0367 & 1.4739 & 1.9763 & 0.2405 & 1.0969 & 2.9188 & 0.3043 & 0.8108 & 3.0485 & 0.4004 \\
Other Specialty Stores & 11 & 0.0059 & -0.1341 & -0.0301 & 0.1519 & 0.3229 & 1.1175 & 0.2370 & 0.0706 & 0.4060 & 0.3223 \\
Investment Managers & 18 & 0.0376 & 0.7707 & 2.4018 & 0.1139 & 0.6134 & 3.1095 & 0.1526 & 0.5251 & 3.1464 & 0.2086 \\
Electrical Products & 13 & 0.0253 & 0.5033 & 1.4703 & 0.1097 & 0.1957 & 1.1061 & 0.1695 & 0.3724 & 2.1064 & 0.2651 \\
Major Pharmaceuticals & 18 & 0.0117 & -0.1838 & -0.0680 & 0.1772 & -0.1492 & -0.3752 & 0.2199 & 0.0132 & 0.1763 & 0.2236 \\
Other Pharmaceuticals & 5 & -0.0034 & 0.3746 & 1.4293 & 0.2405 & 0.1998 & 1.1667 & 0.4583 & 0.1719 & 1.0601 & 1.0000 \\
\bottomrule
\end{tabular}%
}
\caption{Complete result table for Rankformer on NYSE.}
\label{tab:rankformer-training-results-nyse}
\end{table*}

% ======================= TSPRank local stock result ==============================
\begin{table*}[htbp]
\centering
\resizebox{\textwidth}{!}{%
\begin{tabular}{llrrrrrrrrrr}
\toprule
Sector & Stocks &  $\tau$ & IRR@1 & SR@1 & MAP@1 & IRR@3 & SR@3 & MAP@3 & IRR@5 & SR@5 & MAP@5 \\
\midrule
Computer Manufacturing & 5 & 0.0345 & 0.1128 & 0.6215 & 0.1983 & 0.3465 & 2.3286 & 0.5059 & 0.2167 & 1.3724 & 1.0000 \\
Transportation Services & 5 & 0.0008 & 0.4351 & 1.2510 & 0.2489 & 0.2339 & 1.1572 & 0.4838 & 0.1263 & 0.7757 & 1.0000 \\
Medical/Dental Instruments & 17 & 0.0090 & 0.5775 & 1.9145 & 0.0802 & 0.4503 & 2.3065 & 0.1254 & 0.4129 & 2.3442 & 0.1809 \\
Property-Casualty Insurers & 19 & 0.0194 & -0.0452 & 0.1845 & 0.1013 & 0.0247 & 0.2305 & 0.1751 & 0.0401 & 0.2988 & 0.1967 \\
Medical Specialities & 5 & 0.0303 & -0.0239 & 0.1681 & 0.2405 & 0.0473 & 0.3352 & 0.4930 & -0.0402 & -0.1500 & 1.0000 \\
Biotechnology: Biological Products (No Diagnostic Substances) & 12 & 0.0186 & 2.4685 & 2.9704 & 0.1688 & 1.0795 & 3.2313 & 0.2117 & 0.4928 & 1.8969 & 0.3013 \\
Telecommunications Equipment & 16 & 0.0312 & 0.2872 & 0.9428 & 0.0675 & 0.0301 & 0.2521 & 0.1292 & 0.1160 & 0.6588 & 0.2023 \\
\bottomrule
\end{tabular}%
}
\caption{Complete result table for TSPRank with local training on NASDAQ.}
\label{tab:local-training-results}
\end{table*}

\begin{table*}[htbp]
\centering
\resizebox{0.9\textwidth}{!}{%
\begin{tabular}{llrrrrrrrrrr}
\toprule
Sector & Stocks &  $\tau$ & IRR@1 & SR@1 & MAP@1 & IRR@3 & SR@3 & MAP@3 & IRR@5 & SR@5 & MAP@5 \\
\midrule
Biotechnology: Laboratory Analytical Instruments & 5 & 0.0403 & 0.5598 & 2.6040 & 0.2236 & 0.5557 & 3.3714 & 0.5270 & 0.4553 & 3.1498 & 1.0000 \\
Diversified Commercial Services & 9 & 0.0411 & 2.4720 & 2.5039 & 0.2954 & 1.0517 & 3.0297 & 0.3104 & 0.8197 & 3.1495 & 0.4093 \\
Other Specialty Stores & 11 & 0.0227 & 0.0245 & 0.2376 & 0.0717 & 0.6322 & 1.8056 & 0.2046 & 0.4220 & 1.6928 & 0.3128 \\
Investment Managers & 18 & 0.0117 & 1.0477 & 3.4502 & 0.0886 & 0.6783 & 3.7383 & 0.1346 & 0.3774 & 2.6299 & 0.1689 \\
Electrical Products & 13 & 0.0218 & 0.4920 & 1.4221 & 0.1181 & 0.2664 & 1.3944 & 0.1587 & 0.3098 & 1.7686 & 0.2540 \\
Major Pharmaceuticals & 18 & 0.0306 & -0.1847 & -0.0638 & 0.1814 & 0.3830 & 1.4800 & 0.1917 & 0.2538 & 1.4076 & 0.2024 \\
Other Pharmaceuticals & 5 & 0.0092 & 0.4138 & 1.5458 & 0.2405 & 0.3732 & 1.9308 & 0.4695 & 0.1719 & 1.0601 & 1.0000 \\
\bottomrule
\end{tabular}%
}
\caption{Complete result table for TSPRank with local training on NYSE.}
\label{tab:local-training-results-nyse}
\end{table*}

% ======================= TSPRank global stock result ==============================
\begin{table*}[htbp]
\centering
\resizebox{\textwidth}{!}{%
\begin{tabular}{llrrrrrrrrrr}
\toprule
Sector                         & Stocks & $\tau$    & IRR@1   & SR@1    & MAP@1  & IRR@3   & SR@3   & MAP@3  & IRR@5   & SR@5    & MAP@5  \\ \midrule
Computer Manufacturing         & 5      & 0.0462 & 0.7131  & 2.6391  & 0.1941 & 0.4412  & 2.6858 & 0.5063 & 0.2167  & 1.3724  & 1.0000 \\
Transportation Services        & 5      & 0.0790 & 0.8231  & 2.6376  & 0.1772 & 0.5290  & 2.1863 & 0.5211 & 0.1263  & 0.7757  & 1.0000 \\
Medical/Dental Instruments     & 17     & 0.0009 & 0.4891  & 1.3264  & 0.0675 & 0.9051  & 3.1209 & 0.1395 & 0.4674  & 2.3745  & 0.1898 \\
Property-Casualty Insurers     & 19     & 0.0262 & 0.5825  & 1.7370  & 0.1013 & 0.1596  & 0.8014 & 0.1324 & 0.1295  & 0.7147  & 0.1896 \\
Medical Specialities           & 5      & 0.0513 & 0.3184  & 1.1598  & 0.1814 & 0.0501  & 0.3507 & 0.5319 & -0.0402 & -0.1500 & 1.0000 \\
Biotechnology: Biological Products (No Diagnostic Substances) & 12 & 0.0339 & 2.7414 & 3.5486 & 0.1435 & 1.0576 & 3.2959 & 0.2161 & 0.7736 & 2.9855 & 0.3079 \\
Telecommunications Equipment   & 16     & 0.0361 & 0.7213  & 1.6043  & 0.0928 & 0.6366  & 2.5527 & 0.1320 & 0.2411  & 1.3507  & 0.1929 \\
\bottomrule
\end{tabular}%
}
\caption{Complete result table for TSPRank with global training on NASDAQ.}
\label{tab:global-training-results-nasdaq}
\end{table*}

\begin{table*}[htbp]
\centering
\resizebox{0.9\textwidth}{!}{%
\begin{tabular}{llrrrrrrrrrr}
\toprule
Sector                         & Stocks & $\tau$    & IRR@1   & SR@1    & MAP@1  & IRR@3   & SR@3   & MAP@3  & IRR@5   & SR@5    & MAP@5  \\ \midrule
Biotechnology: Laboratory Analytical Instruments & 5 & 0.0454 & 0.5059 & 2.8110 & 0.1983 & 0.4805 & 3.2350 & 0.5251 & 0.4553 & 3.1498 & 1.0000 \\
Diversified Commercial Services & 9 & 0.0287 & 0.6134 & 1.6824 & 0.1603 & 0.7515 & 2.5416 & 0.2771 & 0.5597 & 2.3811 & 0.4168 \\
Other Specialty Stores & 11 & 0.0426 & 0.4883 & 1.1274 & 0.1266 & 0.7789 & 2.1611 & 0.2180 & 0.4915 & 1.7811 & 0.3265 \\
Investment Managers & 18 & 0.0319 & 0.8985 & 3.0370 & 0.0844 & 0.5710 & 2.9765 & 0.1282 & 0.5072 & 3.0107 & 0.1797 \\
Electrical Products & 13 & 0.0364 & 0.5099 & 1.8447 & 0.1097 & 0.3411 & 1.9502 & 0.1793 & 0.3041 & 1.8406 & 0.2766 \\
Major Pharmaceuticals & 18 & 0.0484 & 0.3468 & 0.8992 & 0.1392 & 0.5776 & 2.5628 & 0.1620 & 0.3553 & 2.0901 & 0.2052 \\
Other Pharmaceuticals & 5 & 0.0462 & 0.4623 & 1.6864 & 0.2743 & 0.3309 & 1.6992 & 0.4883 & 0.1719 & 1.0601 & 1.0000 \\
\bottomrule
\end{tabular}%
}
\caption{Complete result table for TSPRank with global training on NYSE.}
\label{tab:global-training-results-nyse}
\end{table*}

\section{MQ2008 Complete Results}
\label{appendix:mq2008-complete}

Table~\ref{tab:mq2008-complete} presents the detailed performance results on the MQ2008 dataset across different folds.

\begin{table*}[htbp]
\centering
\resizebox{\textwidth}{!}{%
\begin{tabular}{l|ccccc|ccccc|ccccc|ccccc}
\toprule
 &
  \multicolumn{5}{c|}{LambdaMART} &
  \multicolumn{5}{c|}{Rankformer} &
  \multicolumn{5}{c|}{TSPRank-Local} &
  \multicolumn{5}{c}{TSPRank-Global} \\ \midrule
 &
  NDCG@3 &
  NDCG@5 &
  NDCG@10 &
  MRR &
  Tau &
  NDCG@3 &
  NDCG@5 &
  NDCG@10 &
  MRR &
  $\tau$ &
  NDCG@3 &
  NDCG@5 &
  NDCG@10 &
  MRR &
  Tau &
  NDCG@3 &
  NDCG@5 &
  NDCG@10 &
  MRR &
  $\tau$ \\ \midrule
 Fold & \multicolumn{20}{c}{Top 10} \\ \midrule
Fold 1 &
  0.6867 &
  0.7180 &
  0.8702 &
  0.4060 &
  0.1330 &
  0.7218 &
  0.7504 &
  0.8832 &
  0.4280 &
  0.2083 &
  0.6837 &
  0.7227 &
  0.8699 &
  0.3956 &
  0.1346 &
  0.7224 &
  0.7606 &
  0.8883 &
  0.4895 &
  0.2174 \\
Fold 2 &
  0.6651 &
  0.7052 &
  0.8631 &
  0.4379 &
  0.1117 &
  0.7057 &
  0.7469 &
  0.8809 &
  0.4784 &
  0.2045 &
  0.6671 &
  0.7031 &
  0.8668 &
  0.4655 &
  0.1193 &
  0.7162 &
  0.7472 &
  0.8838 &
  0.4706 &
  0.1940 \\
Fold 3 &
  0.6904 &
  0.7337 &
  0.8755 &
  0.4286 &
  0.1782 &
  0.7442 &
  0.7743 &
  0.8968 &
  0.4984 &
  0.2651 &
  0.6936 &
  0.7326 &
  0.8735 &
  0.4153 &
  0.1813 &
  0.7341 &
  0.7678 &
  0.8927 &
  0.4883 &
  0.2444 \\
Fold 4 &
  0.6702 &
  0.7180 &
  0.8657 &
  0.4248 &
  0.1406 &
  0.7082 &
  0.7504 &
  0.8814 &
  0.4413 &
  0.2345 &
  0.6930 &
  0.7247 &
  0.8749 &
  0.4662 &
  0.1655 &
  0.7227 &
  0.7519 &
  0.8847 &
  0.4932 &
  0.2192 \\
Fold 5 &
  0.7043 &
  0.7361 &
  0.8791 &
  0.4324 &
  0.1734 &
  0.7301 &
  0.7605 &
  0.8904 &
  0.4844 &
  0.2459 &
  0.6915 &
  0.7232 &
  0.8742 &
  0.3905 &
  0.1714 &
  0.7483 &
  0.7651 &
  0.8926 &
  0.4887 &
  0.2311 \\
\bottomrule
\end{tabular}%
}
\caption{Complete results on MQ2008-list for information retrieval.}
\label{tab:mq2008-complete}
\end{table*}

\section{OTD2 Complete Results}
\label{appendix:events-complete}

Table~\ref{tab:historical-events-complete} shows the detailed performance results on the OTD2 dataset with five different random group allocations.

\begin{table*}[htbp]
\centering
\resizebox{\textwidth}{!}{%
\begin{tabular}{l|cccccccccccccccc}
\toprule
 & \multicolumn{4}{c|}{LambdaMART} & \multicolumn{4}{c|}{Rankformer} & \multicolumn{4}{c|}{TSPRank-Local} & \multicolumn{4}{c}{TSPRank-Global} \\ \midrule
 & \multicolumn{16}{c}{Group Size 10} \\ \midrule
Allocation & $\tau$ $\uparrow$ & EM $\uparrow$ & MRR $\uparrow$ & \multicolumn{1}{c|}{RMSE $\downarrow$} & $\tau$ $\uparrow$ & EM $\uparrow$ & MRR $\uparrow$ & \multicolumn{1}{c|}{RMSE $\downarrow$} & $\tau$ $\uparrow$ & EM $\uparrow$ & MRR $\uparrow$ & \multicolumn{1}{c|}{RMSE $\downarrow$} & $\tau$ $\uparrow$ & EM $\uparrow$ & MRR $\uparrow$ & RMSE $\downarrow$ \\ \midrule
1 & 0.6456 & 0.3043 & 0.7813 & \multicolumn{1}{c|}{1.904} & 0.6269 & 0.2801 & 0.7542 & \multicolumn{1}{c|}{1.962} & 0.5632 & 0.2909 & 0.7703 & \multicolumn{1}{c|}{2.369} & 0.6285 & 0.3275 & 0.8038 & 2.060 \\
2 & 0.5834 & 0.2739 & 0.7504 & \multicolumn{1}{c|}{2.144} & 0.6196 & 0.2924 & 0.7509 & \multicolumn{1}{c|}{2.007} & 0.5664 & 0.2682 & 0.7918 & \multicolumn{1}{c|}{2.226} & 0.6344 & 0.3400 & 0.7997 & 2.047 \\
3 & 0.6343 & 0.3018 & 0.7534 & \multicolumn{1}{c|}{1.963} & 0.6144 & 0.2874 & 0.7162 & \multicolumn{1}{c|}{2.017} & 0.5568 & 0.2917 & 0.7489 & \multicolumn{1}{c|}{2.356} & 0.6304 & 0.3339 & 0.7751 & 2.063 \\
4 & 0.6252 & 0.3051 & 0.7431 & \multicolumn{1}{c|}{1.992} & 0.6123 & 0.2935 & 0.7264 & \multicolumn{1}{c|}{2.022} & 0.5533 & 0.2632 & 0.7363 & \multicolumn{1}{c|}{2.274} & 0.6226 & 0.3310 & 0.7903 & 2.082 \\
5 & 0.6008 & 0.2862 & 0.7538 & \multicolumn{1}{c|}{2.086} & 0.6217 & 0.2962 & 0.7328 & \multicolumn{1}{c|}{1.984} & 0.5891 & 0.3138 & 0.7924 & \multicolumn{1}{c|}{2.255} & 0.6347 & 0.3428 & 0.7993 & 2.035 \\
\bottomrule
\end{tabular}%
}
\caption{Complete results on OTD2 for historical events ordering.}
\label{tab:historical-events-complete}
\end{table*}

\end{document}